\documentclass{article}
\usepackage{graphicx}
\usepackage{xcolor}
\usepackage{booktabs}
\definecolor{lightgray}{gray}{0.92}

\usepackage[final]{neurips_2025}

\usepackage{xspace}
\newcommand{\method}{HypLoRA\xspace}
\usepackage[utf8]{inputenc} 
\usepackage[T1]{fontenc}    
\usepackage{url}            
\usepackage{booktabs}       
\usepackage{amsfonts}       
\usepackage{nicefrac}       
\usepackage{microtype}      
\usepackage{xcolor}         
\usepackage{hyperref} 
\definecolor{darkblue}{rgb}{0.0, 0.0, 0.55} 
\definecolor{darkred}{rgb}{0.55, 0.0, 0.0} 
\hypersetup{
    colorlinks=true, 
    citecolor=darkblue, 
    linkcolor=darkred, 
    urlcolor=darkred 
}
\usepackage{multirow}
\usepackage{tcolorbox}
\usepackage{colortbl}
\usepackage{amsmath}
\usepackage{amsthm} 
\usepackage{graphicx} 
\usepackage{caption}
\usepackage{titletoc}
\usepackage{comment}
\newtheorem{remark}{Remark} 
\newtheorem{proposition}{Proposition}

\definecolor{easycolor}{RGB}{176,196,222}
\definecolor{hardcolor}{RGB}{230,241,243}
\definecolor{mediumcolor}{RGB}{203, 219, 233}
\definecolor{LightGray}{gray}{0.8}

\usepackage{tabularx}
\definecolor{grey}{RGB}{128,128,128}
\newtcolorbox{infobox}[1][]
{
  colframe = grey!55,
  colback  = grey!10,
  #1 
}

\title{Hyperbolic Fine-Tuning for Large Language Models}
\author{
\textbf{Menglin Yang$^{1,2}$, Ram Samarth B B$^3$, Aosong Feng$^4$, Bo Xiong$^5$} \\
\textbf{Jiahong Liu$^6$, Irwin King$^6$, Rex Ying$^4$} \\
${^1}$HKUST(GZ); ${^2}$HKUST; ${^3}$Indian Institute of Science; \\ ${^4}$Yale University; ${^5}$Stanford University; ${^6}$The Chinese University of Hong Kong \\
\texttt{menglin.yang@outlook.com,ramsamarthbb@iisc.ac.in, aosong.feng@yale.edu,} \\ \texttt{xiongbo@stanford.edu, \{jhliu22, king\}@cse.cuhk.edu.hk, rex.ying@yale.edu}  \\
Code:~\url{https://github.com/marlin-codes/HypLoRA} \\ 
Project\thanks{Corresponding author: \texttt{Menglin Yang}. 
}: ~\url{https://hyperboliclearning.github.io/work/hyplora} \\
}

\begin{document}

\maketitle

\begin{abstract}
Large language models (LLMs) have demonstrated remarkable performance across various tasks. However, it remains an open question whether the default Euclidean space is the most suitable choice for LLMs.
In this study, we investigate the geometric characteristics of LLMs, focusing specifically on tokens and their embeddings.
Our findings reveal that token frequency follows a power-law distribution, where high-frequency tokens (e.g., ``the,'' ``that'') constitute the minority, while low-frequency tokens (e.g., ``apple,'' ``dog'') constitute the majority. Furthermore, high-frequency tokens cluster near the origin, whereas low-frequency tokens are positioned farther away in the embedding space.
Additionally, token embeddings exhibit hyperbolic characteristics, indicating a latent tree-like structure within the embedding space.
Motivated by these observations, we propose \textbf{\method}, an efficient fine-tuning approach that operates in hyperbolic space to exploit these underlying hierarchical structures better.
\textbf{\method} performs low-rank adaptation directly in hyperbolic space, thereby preserving hyperbolic modeling capabilities throughout the fine-tuning process.
Extensive experiments across various base models and reasoning benchmarks, specifically arithmetic and commonsense reasoning tasks, demonstrate that \method{} substantially improves LLM performance.
\end{abstract}
\section{Introduction}
\label{sec:1_intro}
   Large language models (LLMs) such as GPT-4~\citep{achiam2023gpt}, LLaMA~\citep{touvron2023llama}, Gemma~\citep{google2024gemma}, and Qwen~\citep{yang2024qwen2} have demonstrated remarkable capabilities in understanding and generating human-like text~\citep{qin2023chatgpt,shen2024hugginggpt,liu2025survey}.
   Despite their impressive capabilities, these models often rely on Euclidean geometry for token representation, which may inadequately capture the inherently complex and hierarchical nature of real-world data structures~\citep{bronstein2017geometric,bachmann2020constant,suzuki2021generalization,nickel2017poincare,krioukov2010hyperbolic,sarkar2011low}.
    Consider how words naturally organize into nested categories with varying levels of abstraction: abstract concepts like ``fruit'' occupy higher positions in the semantic hierarchy, while specific instances such as ``apple'' or ``banana'' populate the lower levels.
    Representing such structures effectively is crucial for understanding the semantics of language in LLMs.

Recent advancements suggest that non-Euclidean geometries, particularly hyperbolic spaces~\citep{nickel2017poincare,nickel2018learning,ganea2018hyperbolic,khrulkov2020hyperbolic,cetin2022hyperbolic,peng2021hyperbolic,yang2022hyperbolic,liu2024client,mettes2023hyperbolic,yang2022hrcf,yang2022hicf}, offer promising alternatives for modeling hierarchical data. Hyperbolic space, distinguished by its negative curvature, is especially well-suited for representing tree-like hierarchical data due to its exponential volume growth and geometric prior. 
This geometric property makes hyperbolic space particularly capable for tasks involving complex, hierarchically structured information. 

\textbf{Proposed Analysis Framework.}
In this work, we first delve into how LLMs interact with token embeddings and explore the extent to which these embeddings exhibit non-Euclidean characteristics. We approach this from both a \texttt{global} and \texttt{local} perspective.
   At the \texttt{global} level, we analyze the overall distribution of tokens by frequency and investigate how these frequencies are arranged across the embedding space. At the \texttt{local} level, we measure the hyperbolicity of the metric space spanned by each input prompt, where the hyperbolicity 
   serves as a proxy for evaluating the distance or dissimilarity between the underlying embedding structure and a tree-like hierarchy~\citep{borassi2015hyperbolicity,kennedy2013hyperbolicity,khrulkov2020hyperbolic}.

\begin{figure}[t]
   \centering
    \includegraphics[width=0.240\textwidth]{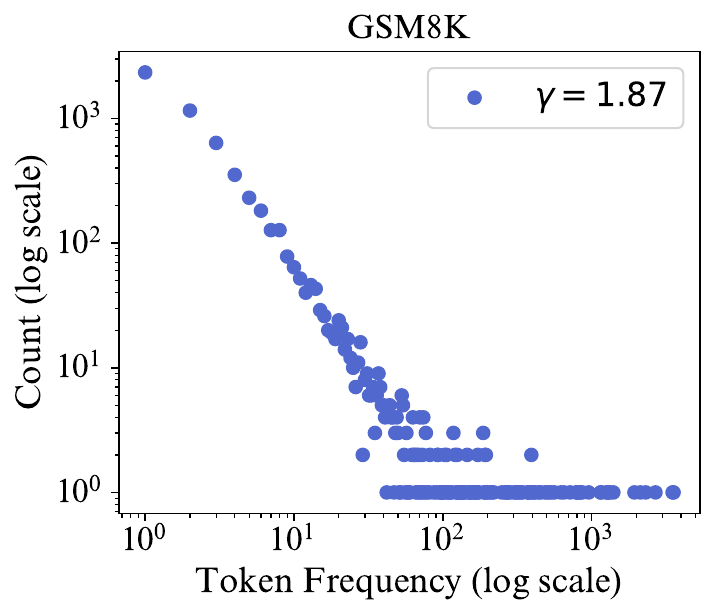}
    \includegraphics[width=0.240\textwidth]{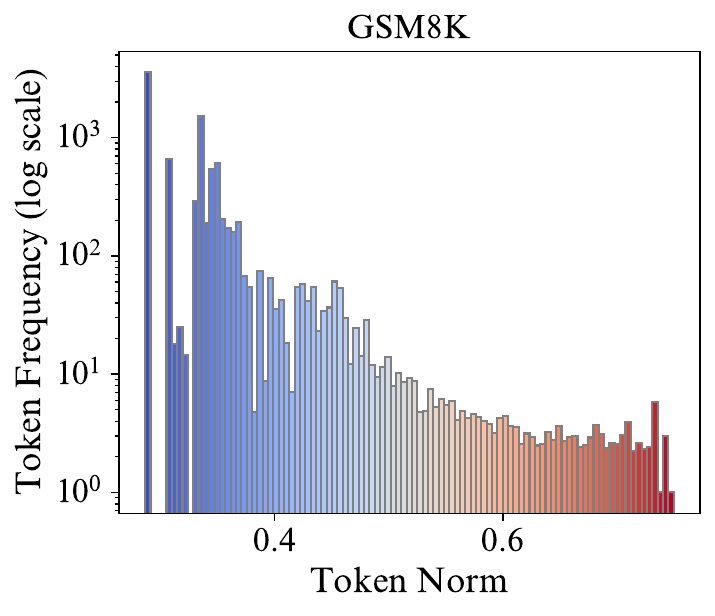}
    \includegraphics[width=0.240\textwidth]{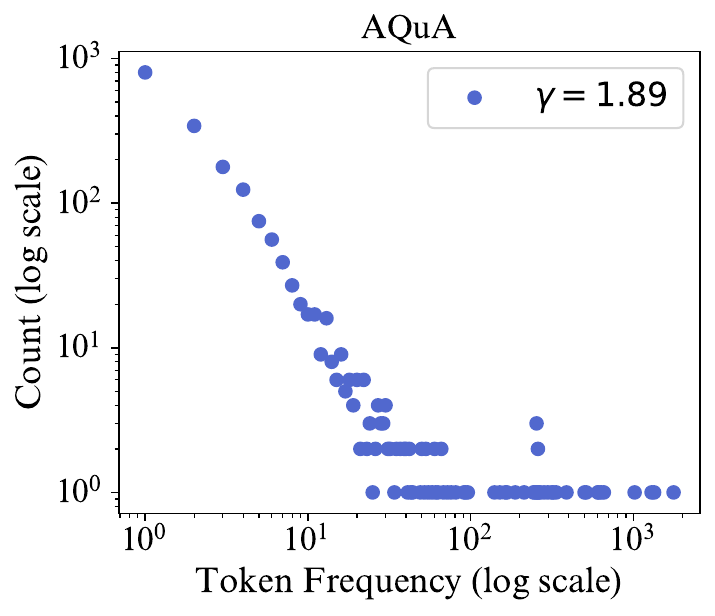}
    \includegraphics[width=0.240\textwidth]{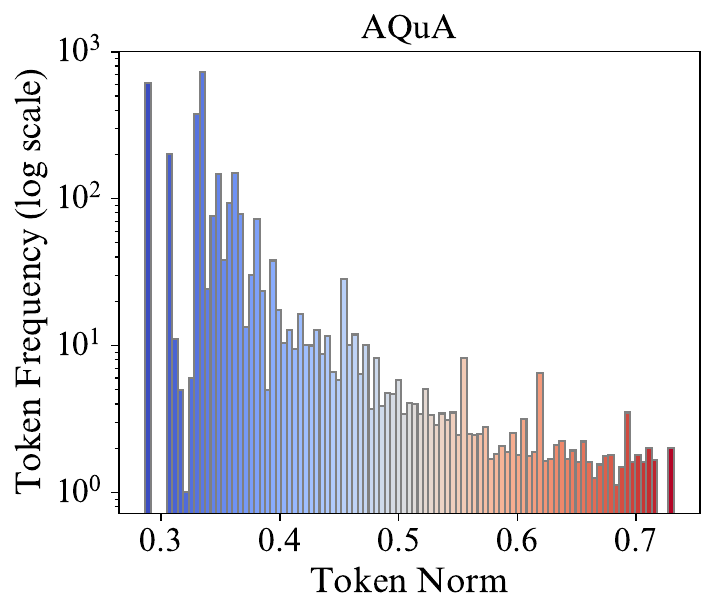}
   \caption{Token frequency distribution and token frequency vs.\ norm analysis for GSM8K (Group 1) and AQuA (Group 2) datasets in LLaMA3-8B. For each group, the left panels show the token frequency distributions (power-law distribution), while the right panels illustrate the relationship between token frequency and the corresponding norms. This visualization reveals the underlying geometric structure of the token embeddings. For additional data analysis and visualizations, please refer to Appendix~\ref{sec:appendix_more_investigation}.}
   \label{fig:token_frequency_distribution}
\end{figure}

    Our analysis in Section~\ref{sec:hyperbolicity_investigation} reveals several key insights. 
    \texttt{Globally}, token frequency follows a power-law distribution, where high-frequency tokens (e.g., ``the,'' ``that'') constitute the minority, while low-frequency tokens (e.g., ``apple,'' ``dog'') constitute the majority. 
    Power-law distributions are consistent with, and can naturally arise from, underlying 
    hierarchical or branching generative 
mechanisms~\citep{krioukov2010hyperbolic,barabasi2003scale,alvarez2006hierarchical}.\footnote{Power-law scaling alone does not uniquely identify the underlying structure 
or mechanism; additional evidence is needed to support a hierarchical 
interpretation~\citep{nakaishi2025rethinking}.}
    Besides, high-frequency tokens (e.g., abstract concepts, function words) tend to be located near the origin of the embedding space, while low-frequency tokens (e.g., specific terms) are farther away, as demonstrated in Table~\ref{tab:tokens_norm}. \texttt{Locally}, our investigation of hyperbolicity ($\delta$ values) in Table~\ref{tab:hyperbolicity_table}
 demonstrates that LLM token embeddings in each prompt exhibit significant tree-like properties. 

Based on our findings above, a natural consideration is to develop hyperbolic LLMs that explicitly incorporate a hyperbolic inductive bias\footnote{The connection between power-law distribution and hyperbolic geometry is elaborated in Section~\ref{sec:connection_powerlaw_hyperbolic}.}. 
However, training LLMs from scratch is resource-intensive~\citep{loshchilov2017decoupled,rajbhandari2020zero,hu2021lora}. 
As a more resource-efficient alternative, we propose to build the first low-rank adaptation fine-tuning method in hyperbolic space. 
This approach is particularly advantageous given that existing LLMs are all Euclidean, and not all downstream tasks require hyperbolic geometry in their fine-tuning. {By employing hyperbolic adapters on Euclidean LLMs for specific tasks, we can leverage the benefits of both geometries while maintaining computational efficiency}.

\textbf{Challenges.} Adapting LLMs in non-Euclidean embedding spaces with classic techniques, \emph{i.e.}, applying exponential and logarithmic maps within tangent space~\citep{chami2019hyperbolic,ganea2018hyperbolic_hnn,yang2022hyperbolichtgn,cho2023curve,fu2024hyperbolic} for weight adaptation is problematic in this case. This approach fails to fully capture the hyperbolic geometry, as the exponential and logarithmic maps are mutually inverse and can be canceled with {consecutive operations}\footnote{The cancellation effect occurs because standard hyperbolic neural network~\cite{ganea2018hyperbolic_hnn,chami2019hyperbolic} approaches apply transformations in the tangent space at the origin, requiring the sequence: Euclidean embedding → exponential map to hyperbolic space → logarithmic map to tangent space → linear transformation → exponential map back to hyperbolic space → projection to Euclidean space.
When these operations are chained together, the maps are mutually inverse and effectively cancel out, reducing the entire sequence to approximately the original Euclidean transformation BAx without preserving the beneficial hyperbolic geometry.}. Consequently, the inherent properties of the hyperbolic space are not effectively preserved, limiting the potential benefits of incorporating non-Euclidean geometries into the adaptation process.

\textbf{Proposed Method.}
To address this limitation, we introduce \textbf{\method} to perform low-rank adaptation directly on the hyperbolic manifold without transformation to the tangent space, thus preserving hyperbolic modeling capabilities and counteracting the cancellation effect. \method integrates hyperbolic geometry into existing LLMs, implicitly introducing high-order interactions and accounting for token hierarchies, enabling them to benefit from hyperbolic characteristics while minimizing additional computational costs. 

To summarize, our main contributions are twofold: (1) We conduct a comprehensive investigation into the geometric characteristics of token embeddings in LLMs, revealing their inherent tree-like structure and strong hyperbolic properties. (2) We propose \method, a parameter-efficient fine-tuning method that integrates hyperbolic geometry into LLMs while keeping it aligned with the Euclidean LLM framework. 
We conduct extensive experiments on various models and different tasks, specifically arithmetic reasoning and commonsense reasoning, demonstrating clear advantages over competitive baselines.

\section{Related Work}

\textbf{Hyperbolic Representation Learning and Foundation Models.} 
Hyperbolic geometry has been successfully applied to various neural network architectures and models \citep{yang2022hyperbolic,mettes2023hyperbolic,peng2021hyperbolic}, including shallow hyperbolic neural networks \citep{ganea2018hyperbolic,ganea2018hyperbolic_hnn,chen2021fully,shimizu2020hyperbolic,fan2025curvature,mao2024klein}, hyperbolic CNNs \citep{bdeir2023hyperbolic,van2023poincare,khan2025hyperbolic}, hyperbolic GNNs~\citep{chami2019hyperbolic,liu2019hyperbolic,zhang2021hyperbolic,yang2023kappahgcn}, and hyperbolic attention networks or Transformers~\citep{gulcehre2018hyperbolic,chen2021fully,shimizu2020hyperbolic,yang2024hypformer}. These models leverage the inductive biases of hyperbolic geometry to achieve remarkable performance on various tasks and applications~\citep{chami2019hyperbolic,yang2022hrcf,yang2022hicf,sun2021hgcf,khrulkov2020hyperbolic,cetin2022hyperbolic,weng2021unsupervised,xiong2022hyperbolic,yang2021discrete,gao2021curvature}.
Recent efforts have focused on adapting LLMs and CLIP~\citep{radford2021learning} to hyperbolic spaces. Key advancements include developing more expressive hyperbolic image-text representations~\citep{desai2023meru}, enabling compositional entailment learning for deeper vision-language understanding~\citep{pal2024compositional}, designing safety-aware hyperbolic frameworks for content moderation~\citep{poppi2025hyperbolic}, and creating core modules to facilitate the construction of novel hyperbolic foundation models~\citep{he2025hypercore}. 
While these adaptations show promise, training LLMs from scratch remains computationally expensive \citep{kochurov2020geoopt,smith2014optimization}. 
The computational complexity increases further when considering Riemannian optimization \citep{kochurov2020geoopt,smith2014optimization,becigneul2018riemannian} and additional hyperbolic operations, like M\"obius addition. 

\textbf{Geometric Analysis of Language Model Embeddings.} Prior work has made important observations about the geometry of embeddings that helped shape and motivate our research. Reif et al.~\citep{reif2019visualizing} demonstrated that BERT embeddings contain distinct syntactic and semantic subspaces and showed evidence of tree-like parse structures, while Gao et al.~\citep{gao2019representation} revealed that token embeddings tend to cluster in a narrow cone during training, leading to representation degeneration. Building on these geometric insights, Rudman et al.~\citep{rudman2021isoscore} introduced IsoScore to formally quantify how uniformly embeddings utilize the ambient vector space. Additionally, Puccetti et al.~\citep{puccetti2022outliers} analyzed outlier dimensions in Transformers and showed their correlation with token frequencies. 
While these works provide crucial foundations for understanding embedding geometry, our work differs in that we specifically quantify and leverage the natural hyperbolicity of token embeddings.

\textbf{Parameter-Efficient Fine-Tuning (PEFT) and LoRA.} Fine-tuning LLMs~\citep{openai2022, achiam2023gpt, touvron2023llama} for downstream tasks poses significant challenges due to their massive number of parameters.
To address this issue, PEFT methods have been proposed, which aim to train a small subset of parameters while achieving comparable or even better performance compared to full fine-tuning. PEFT methods can be broadly categorized into prompt-based methods~\citep{lester2021power,li2021prefix,qin2021exploring}, adapter-based methods~\citep{houlsby2019parameter,zhu2021counter}, and reparameterization-based methods~\citep{hu2021lora,aghajanyan2020intrinsic,edalati2022krona}.
Among these, the reparameterization-based LoRA~\citep{hu2021lora} has gained significant attention due to its simplicity, effectiveness, and compatibility with existing model architectures. 
Variants of LoRA, such as LoRA+\citep{hayou2024lora+}, DoRA~\citep{liu2024dora}, and AdaLoRA~\citep{zhang2023adaptive}, have been proposed to improve its performance and efficiency. Recent research has also investigated ensembles of multiple LoRAs~\citep{wang2023lora,ren2024mini} and quantization techniques~\citep{dettmers2024qlora,xu2023qa,li2023loftq}. 
The proposed method is a foundational algorithm that is orthogonal to existing approaches and can potentially be combined with various LoRA variants to exploit their complementary strengths and achieve superior performance.

\section{Preliminary}

This section introduces the key concepts used in our study, including the Lorentz model of hyperbolic geometry and the LoRA adapter.

\textbf{Hyperbolic Geometry.} Unlike flat Euclidean geometry, hyperbolic geometry is characterized by a constant negative curvature. We utilize the Lorentz model, also known as the hyperboloid model due to its ability to effectively capture hierarchical structures and maintain numerical stability \citep{nickel2018learning,chen2021fully,mishne2023numerical}. The Lorentz model in $n$ dimensions with curvature $-1/K (K>0)$ is defined as:
\begin{equation}
\mathcal{L}_K^n = \{\mathbf{x} \in \mathbb{R}^{n+1} : \langle \mathbf{x}, \mathbf{x} \rangle_\mathcal{L} = -K, {x}_0 > 0\},
\end{equation}
where $\langle \cdot, \cdot \rangle_\mathcal{L}$ is the Lorentzian inner product, given by:
$
\langle \mathbf{x}, \mathbf{y} \rangle_\mathcal{L} = -{x}_0 {y}_0 + \sum_{i=1}^n {x}_i {y}_i.
$

\textbf{Tangent Space.} In the Lorentz model $\mathcal{L}_K^n$, the tangent space at a point $\mathbf{x}$ is denoted  $\mathcal{T}_{\mathbf{x}} \mathcal{L}_K^n$. It is defined as the set of all vectors $\mathbf{u}$ that are orthogonal to $\mathbf{x}$ under the Lorentzian inner product:
\begin{equation}
\mathcal{T}_{\mathbf{x}} \mathcal{L}_K^n := \{\mathbf{u} \in \mathbb{R}^{n+1} : \langle \mathbf{u}, \mathbf{x} \rangle_{\mathcal{L}} = 0\}.
\end{equation}
To facilitate projection between the hyperboloid and its tangent spaces at $\mathbf{x}$, one can utilize two critical mappings: the exponential and logarithmic maps. The \textit{exponential map} at $\mathbf{x}$, denoted $\exp_{\mathbf{x}}^K$, projects a vector from the tangent space $\mathcal{T}_{\mathbf{x}} \mathcal{L}_K^n$ back onto the hyperboloid. Conversely, the \textit{logarithmic map}, denoted $\log_{\mathbf{x}}^K$, maps a point on the hyperboloid to the tangent space at $\mathbf{x}$.
The detailed formulas are given in Appendix~\ref{appendix:exponential_and_logarithmic_map}.

\textbf{LoRA Adapter.} The LoRA adapter provides an efficient approach for modifying LLMs with minimal computational overhead. Instead of retraining the entire model, LoRA focuses on adjusting specific components within the model's architecture to transform an input $\mathbf{x}\in\mathbb{R}^d$ into an output $\mathbf{z}\in\mathbb{R}^k$.
In practice, LoRA targets the weight matrices found in each Transformer layer of an LLM. Typically, the weight $W$ of the Transformer, which resides in the dimensions $\mathbb{R}^{k \times d}$, is adapted through a low-rank approximation. This is achieved by introducing an additional term, $\Delta W$, to the original weight matrix:
\begin{equation}
\begin{aligned}
  \mathbf{z} = W_\text{LoRA}(\mathbf{x})
  = W\mathbf{x} + \Delta W\mathbf{x} = W\mathbf{x} + BA\mathbf{x}. 
  \label{equ:euclidan_lora}
\end{aligned}
\end{equation}
Here, $A \in \mathbb{R}^{r \times d}$ and $B \in \mathbb{R}^{k \times r}$ represent two smaller, learnable matrices where $r$ is the rank of these matrices, which is significantly less than either $d$ or $k$. This design choice ensures that $r \ll \min(d, k)$, thereby reducing the complexity of the model adaptation.
During the fine-tuning process, only the matrices $A$ and $B$ are adjusted, while the pre-existing weights $W$ are kept frozen. This method significantly decreases the number of parameters that need to be trained, from $d\cdot k$ to $(d+k)\cdot r$, enhancing the efficiency of the fine-tuning process. As a result, LoRA enables the targeted adaptation of LLMs, allowing them to transform an input $\mathbf{x}$ into an output $\mathbf{z}$ while maintaining high performance and adapting to new tasks or datasets with a fraction of the computational resources typically required.

\section{Investigation}
\label{sec:hyperbolicity_investigation}
In this section, we present an in-depth investigation of token embeddings in LLMs from both global and local perspectives. Our goal is to uncover the geometric structures underlying pretrained token representations, specifically examining the global distribution of token frequencies and their spatial arrangement, as well as the local hyperbolicity of token embeddings across various datasets.

\subsection{Global Token Statistics}
\label{sec:global}
We begin by investigating the global distribution of token frequencies in the context of arithmetic reasoning datasets, focusing on datasets such as GSM8K~\citep{cobbe2021training}, AQuA~\citep{ling2017program}, MAWPS~\citep{koncel2016MAWPS}, and SVAMP~\citep{patel2021nlp}. We also provide a broader analysis across different types of datasets and LLMs in Appendix~\ref{sec:appendix_more_investigation}.
Figure~\ref{fig:token_frequency_distribution} (left) presents the distribution of token frequencies, with a power-law exponent $\gamma \approx 1.9$, as estimated by the \texttt{powerlaw} package~\citep{alstott2014powerlaw}. In such distributions, the exponent $\gamma$ controls how quickly token frequencies decline: smaller values of $\gamma$ (closer to 1) indicate a more gradual decay where frequent tokens dominate, while larger values signify a sharper decline, with most tokens being rare.

This power-law behavior is consistent with the tree-like hierarchical nature of language~\citep{nickel2017poincare,khrulkov2020hyperbolic,ravasz2003hierarchical,yang2021discrete,yang2022hicf}. High-frequency tokens often correspond to more abstract or general concepts, while low-frequency tokens represent specific or rare terms. This pattern aligns with a hierarchical organization of the token space: abstract, high-frequency tokens cluster 
near the origin, while specific terms are positioned farther out, mirroring how general concepts sit at the core of a semantic hierarchy with specialized terms at the periphery.

\begin{table}[t!]
\centering
\caption{Mean, Minimum, and Maximum frequency and norm values of token embedding in different base models and groups. Group 1: \textit{to, in, have, that, and, is, for}, Group 2: \textit{how, much, many, time, cost}, Group 3: \textit{animal, fruit, number, color, size}, Group 4: \textit{dog, cow, apple, banana, 380, 480, purple, red, medium, small, large}.}
\small
\resizebox{0.75\textwidth}{!}{%
\begin{tabular}{llrr}
\toprule
\textbf{Model}   & \textbf{Group}   & \textbf{Frequency (Mean [Min$\sim$Max])} & \textbf{Norm (Mean [Min$\sim$Max])} \\
\midrule
\multirow{4}{*}{Gemma-7B}  & Group 1 & $4934.4$ [$1838\sim8539$] & $3.160$ [$3.060\sim3.299$] \\
                           & Group 2 & $2709.4$ [$474\sim6681$]  & $3.561$ [$3.488\sim3.627$] \\
                           & Group 3 & $292.0$ [$34\sim1191$]    & $3.765$ [$3.623\sim3.887$] \\
                           & Group 4 & $114.3$ [$25\sim284$]     & $3.998$ [$3.660\sim4.520$] \\
\midrule
\multirow{4}{*}{LLaMA-7B}  & Group 1 & $4993.9$ [$1838\sim8547$] & $0.951$ [$0.793\sim1.060$] \\
                           & Group 2 & $2712.6$ [$474\sim6683$]  & $1.222$ [$1.118\sim1.299$] \\
                           & Group 3 & $299.8$ [$34\sim1200$]    & $1.325$ [$1.274\sim1.428$] \\
                           & Group 4 & $139.1$ [$26\sim286$]     & $1.364$ [$1.326\sim1.417$] \\
\midrule
\multirow{4}{*}{LLaMA3-8B} & Group 1 & $4937.4$ [$1838\sim8547$] & $0.353$ [$0.330\sim0.396$] \\
                           & Group 2 & $2710.0$ [$474\sim6683$]  & $0.456$ [$0.394\sim0.499$] \\
                           & Group 3 & $292.6$ [$34\sim1191$]    & $0.499$ [$0.452\sim0.549$] \\
                           & Group 4 & $97.1$ [$13\sim284$]      & $0.569$ [$0.499\sim0.675$] \\
\midrule
\multirow{4}{*}{LLaMA-13B} & Group 1 & $4993.9$ [$1838\sim8547$] & $1.027$ [$0.833\sim1.255$] \\
                           & Group 2 & $2712.6$ [$474\sim6683$]  & $1.429$ [$1.346\sim1.489$] \\
                           & Group 3 & $299.8$ [$34\sim1200$]    & $1.494$ [$1.453\sim1.532$] \\
                           & Group 4 & $139.1$ [$26\sim286$]     & $1.501$ [$1.470\sim1.526$] \\
\bottomrule
\end{tabular}
}
\label{tab:tokens_norm}
\vspace{-10pt}
\end{table}

\textbf{Empirical Observation.} To better understand the relationship between token frequency and their spatial arrangement within the embedding space, we calculate the average token frequency as a function of their distance from the origin. As shown in Figure~\ref{fig:token_frequency_distribution} (right), 
high-frequency tokens (e.g., ``the,'' ``that'') tend to have smaller norms, 
while low-frequency tokens (e.g., ``apple,'' ``dog'') have larger norms.
Table~\ref{tab:tokens_norm} presents representative tokens across different frequencies 
and norm ranges within the embedding space of different base models.
We categorize tokens into four groups based on their linguistic function and specificity: Group 1 contains high-frequency function words (e.g., \textit{to, is, and}), 
Group 2 contains common question/quantity words (e.g., \textit{how, much, many}), 
Group 3 contains general category nouns (e.g., \textit{animal, fruit, color}), 
and Group 4 contains specific instances (e.g., \textit{dog, apple, purple}).

The results presented in Table~\ref{tab:tokens_norm} demonstrate several critical findings. \textit{First}, we observe a statistically significant separation between functional/abstract words (Group 1) and specific terms (Group 4) across all models, with Group 1 consistently exhibiting the smallest embedding norms and highest frequencies, while Group 4 shows the largest norms and lowest frequencies. \textit{Second}, the relative ordering of groups remains consistent across all examined models, with $\text{Group 1} < \text{Group 2} < \text{Group 3} < \text{Group 4}$ in terms of embedding norms, despite absolute magnitude variations. 
Most notably, even across different architectural families (LLaMA vs. Gemma), the hierarchical organization principle remains preserved, though with different absolute scales, where Gemma-7B exhibits systematically larger embedding norms (mean Group 1 norm: 3.160) compared to LLaMA models (mean Group 1 norm: 0.353 $\sim$ 1.027), yet maintains the same relative hierarchical structure.

{\textbf{Conclusion (1) } These findings suggest that the spatial organization of token embeddings reflects the inherent hierarchical relationships in language, supporting the hypothesis that token embedding in LLMs exhibits a tree-like structure, with spatial positioning aligned with token frequency and specificity.} It is worth noting, however, that a power-law distribution of token frequency alone does not guarantee the emergence of a hierarchical token embedding, as it also depends on the training objectives. 
Our analysis demonstrates that the hierarchy is strongly correlated with token frequencies, which can be understood through the lens of LLMs' tokenization and co-occurrence pattern learning during training~\cite{sennrich2016neural}.
While the exact mechanisms underlying this relationship require further investigation in future work, the spatial distribution of token embeddings remains crucial as it provides the primary motivation for our methodological approach.

\subsection{$\delta$-Hyperbolicity of Local Token Embeddings}
\label{sec:local}
To rigorously quantify the hierarchical nature of token embeddings, we further examine the $\delta$-hyperbolicity of the space spanned by the token embedding. $\delta$-hyperbolicity, introduced by Gromov~\citep{gromov1987hyperbolic}, is a measure that captures the degree to which a metric space deviates from an exact tree structure. Lower values of $\delta$ imply a space more similar to a perfect tree, while higher values indicate deviation from a tree-like structure.

We compute $\delta$-hyperbolicity using the four-point condition, which compares the Gromov products between any four points $a$, $b$, $c$, and $w$ in the metric space. Specifically, the hyperbolicity is defined as:
\begin{equation}
[a, c]_w \geq \min([a, b]_w, [b, c]_w) - \delta,
\end{equation}
where the Gromov product \([a, b]_w\) is:
\begin{equation}
[a, b]_w = \frac{1}{2}(d(a, w) + d(b, w) - d(a, b)).
\end{equation}

\textbf{Quantitative Analysis}. To measure the hyperbolicity of token embeddings, we apply this algorithm to various open-source LLMs. Following the methodologies proposed by Khrulkov et al.~\citep{khrulkov2020hyperbolic} and Cetin et al.~\citep{cetin2022hyperbolic}, we estimate $\delta$-hyperbolicity using the efficient algorithm introduced by Fournier et al.~\citep{fournier2015computing}. To ensure scale invariance, we normalize $\delta$ by the diameter of the embedding space, $\operatorname{diam}(X)$, yielding a relative measure:
$
\delta_{rel} = \frac{2\delta}{\operatorname{diam}(X)}.
$
This relative measure ranges from 0 to 1, with values closer to 0 indicating a highly hyperbolic (tree-like) structure, and values near 1 indicating a non-hyperbolic, flat structure.
We employ Euclidean distance as a measure of the shortest distance, maintaining the same computational paradigm as in previous works~\citep{khrulkov2020hyperbolic, cetin2022hyperbolic}. To further validate the correctness of this approach, we generate a series of random graphs with predefined hyperbolicity, embed them using a two-layer graph neural network (GNN)~\citep{kipf2016semi}, and then compute the hyperbolicity. Details of this process are provided in Appendix~\ref{sec:hyperbolicity_different_metric_space}. Our experiments reveal a positive correlation between the hyperbolicity of the embeddings and the original graphs. Consequently, we utilize this method as a proxy for estimating the hyperbolicity of token embeddings.
In our analysis, we calculate hyperbolicity at the prompt level, treating each token within a prompt as a point in the metric space spanned by the embeddings. By averaging the hyperbolicity across all prompts, we assess the overall hyperbolic structure of token embeddings in each dataset.

{\textbf{Conclusion (2)} Our results, as shown in Table~\ref{tab:hyperbolicity_table}, reveal that token embeddings exhibit significant hyperbolicity, suggesting that the embedding space has a strong tree-like structure.} This observation further corroborates our findings from the global token statistics, where the arrangement of tokens in the embedding space mirrors hierarchical relationships seen in language data. We also provide the hyperbolicity analysis of the final hidden layer in Appendix~\ref{sec:hyperbolicity-final-layer}.

\begin{table*}[!t]
\centering
\caption{Comparison of $\delta$-Hyperbolicity across various metric spaces and datasets. The left table provides reference values for baseline metric spaces, allowing for a clearer interpretation of hyperbolicity in the analyzed datasets in the right table.}
\vspace{2pt}
\begin{tabular}{p{0.32\textwidth} p{0.65\textwidth}}
\centering
\resizebox{0.35\textwidth}{!}{
\begin{tabular}{l c}
\toprule
\textbf{Metric Space} & \textbf{Hyperbolicity($\delta$)} \\
\midrule
Sphere Space   & $0.99 \pm 0.01$ \\
Random Graph    & $0.62 \pm 0.34$ \\
PubMed Graph   & $0.40 \pm 0.45$ \\
Scale-free Graph & $0.00$ \\
Tree Graph     & $0.00$ \\
\bottomrule
\end{tabular}
}
&
\centering
\resizebox{0.6\textwidth}{!}{
\begin{tabular}{@{}lcccc@{}}
\toprule
\textbf{Hyperbolicity}$(\delta)$& \multicolumn{1}{c}{MAWPS} & \multicolumn{1}{c}{SVAMP} & \multicolumn{1}{c}{GSM8K} & \multicolumn{1}{c}{AQuA} \\
\midrule
LLaMA-7B   & $0.08 \pm 0.02$ & $0.09 \pm 0.01$ & $0.10 \pm 0.01$ & $0.10 \pm 0.01$ \\
LLaMA-13B  & $0.08 \pm 0.01$ & $0.09 \pm 0.01$ & $0.09 \pm 0.01$ & $0.10 \pm 0.01$ \\
Gemma-7B   & $0.11 \pm 0.01$ & $0.11 \pm 0.01$ & $0.11 \pm 0.01$ & $0.12 \pm 0.01$ \\
LLaMA3-8B  & $0.06 \pm 0.01$ & $0.07 \pm 0.01$ & $0.07 \pm 0.01$ & $0.08 \pm 0.01$ \\
\midrule
Average    & $0.08 \pm 0.01$ & $0.09 \pm 0.01$ & $0.09 \pm 0.01$ & $0.10 \pm 0.01$ \\
\bottomrule
\end{tabular}
}
\end{tabular}
\label{tab:hyperbolicity_table}
\vspace{-10pt}
\end{table*}

\subsection{Connection between Power-law Distribution and Hyperbolic Geometry}
\label{sec:connection_powerlaw_hyperbolic}

Having established both the global power-law distribution (Section~\ref{sec:global})
and local tree-like geometry (Section~\ref{sec:local}) of token embeddings, we now
examine the theoretical connection between these two observations.

The observation of a power-law distribution in token frequencies, as discussed in Section~\ref{sec:hyperbolicity_investigation}, is not merely a statistical curiosity. It has deep connections to the underlying geometry of the data, particularly to hyperbolic spaces, which are well-suited for representing hierarchical structures~\citep{nickel2017poincare, krioukov2010hyperbolic,krioukov2009curvature,papadopoulos2010greedy}. 
For instance, Nickel and Kiela~\citep{nickel2017poincare} highlighted that the existence of power-law degree distributions can often be traced back to hierarchical structures. 
Similarly, Ravasz and Barabási~\citep{ravasz2003hierarchical} established that the scaling law $P(k) \sim k^{-\gamma}$ can signify the co-existence of a hierarchy of nodes with varying degrees of clustering. 
Krioukov et al.~\citep{krioukov2010hyperbolic} further strengthened this connection by showing that the exponent of the power-law degree distribution is a function of the hyperbolic space curvature. 
Building on this geometric understanding, Papadopoulos et al.~\citep{papadopoulos2010greedy} demonstrated that complex (scale-free) network topologies naturally emerge when networks grow within an underlying hyperbolic metric space, and importantly, that the resulting hyperbolic embedding of these dynamic scale-free networks facilitates highly efficient greedy forwarding.

To formalize this connection with hyperbolic geometry, we can consider embedding tokens in a hyperbolic space. A common model for hyperbolic space is the Poincaré disk model ($\mathbb{H}^2$) with curvature $K = -1$\footnote{The derivation uses the Poincar\'e 
ball model for its intuitive geometric interpretation. However, all models of 
hyperbolic space, including the Poincar\'e ball, the Lorentz (hyperboloid) model, 
and the Klein model, are isometrically equivalent, preserving geodesic distances 
under explicit diffeomorphisms~\citep{cannon1997hyperbolic,ratcliffe2006foundations}. 
Since our method (Section~\ref{sec:HypLoRA_method}) operates in the Lorentz model, the 
theoretical connections established here between power-law distributions and 
hyperbolic geometry remain fully applicable.}. In such a space, both the circumference $C(r)$ and area $A(r)$ of a circle of radius $r$ exhibit exponential growth:
\begin{equation}
C(r) = 2\pi \sinh(r) \sim e^r \quad \text{as } r \to \infty,
\end{equation}
\begin{equation}
    A(r) = 2\pi (\cosh(r) - 1) \sim e^r \quad \text{as } r \to \infty.
\end{equation}
If we consider token embeddings in a hyperbolic space with polar coordinates $(r, \theta)$, where $r \in \mathbb{R}^+$ is the radial coordinate (correlating with token frequency) and $\theta \in [0, 2\pi]$ is the angular coordinate (encoding semantic similarity), the radial distribution of tokens follows $p(r) \sim e^{-\zeta r}$, where $\zeta > 0$ relates to the hyperbolic curvature $K$. The frequency function $k(r)$ for tokens at radius $r$ is then given by $k(r) \sim e^{-r}$.
Given $k(r) \sim e^{-r}$, we have $r \sim -\ln k$, and thus 
$\left| \frac{dr}{dk} \right| \sim k^{-1}$. Combined with the radial distribution 
$p(r) \sim e^{-\zeta r} \sim k^{\zeta}$, this yields:
\begin{equation}
    P(k) \sim p(r) \left| \frac{dr}{dk} \right| \sim k^{\zeta} \cdot k^{-1} \sim k^{-(1-\zeta)}.
\end{equation}
Following the parameterization of Krioukov et al.~\citep{krioukov2010hyperbolic}, 
the power-law exponent $\gamma$ relates to the curvature parameter via 
$\gamma = 2/\zeta + 1$ in the context of complex networks. This relationship 
underscores the theoretical connection between the power-law behavior observed 
in token frequencies and the inherent hyperbolic geometry of the embedding space.
Since hyperbolic models such as the Poincaré ball model and the Lorentz model are isometric, this conclusion can be extended to other hyperbolic models.

Hyperbolic space offers distinct advantages for modeling language hierarchies, especially when addressing the structural and spatial constraints of token co-occurrence:
\textbf{(1) Separation of Low-Frequency Tokens.} Tokens with low frequencies, which typically represent more specific or granular concepts, require clear separation from each other to maintain semantic clarity.
\textbf{(2) Proximity to High-Frequency Hypernyms.} Simultaneously, these low-frequency tokens should remain close to their corresponding high-frequency hypernyms or function words.
Hyperbolic space is uniquely suited for capturing these dual constraints due to its exponential volume growth, which inherently supports hierarchical structure and allows for ample separation of specific entities while keeping them close to their parent categories. This contrasts with Euclidean space, where such arrangements can lead to crowding or distortion of distances.

\textbf{Overall Conclusion.} Through these analyses, we demonstrate that token embeddings in LLMs exhibit hierarchical organization and significant hyperbolicity. This understanding not only sheds light on the geometric nature of token embeddings but also motivates the development of methods that can better capture and preserve these underlying geometric properties.\footnote{While our analysis reveals consistent hierarchical patterns 
across multiple LLMs, several limitations should be noted. First, our investigation 
focuses on arithmetic reasoning and commonsense datasets (please check Appendix~\ref{sec:appendix_more_investigation} for details); the generalizability to other domains 
(e.g., code, multilingual text) requires further validation. Second, the relationship 
between token frequency and embedding norm, while strong, is correlational rather 
than causal. Third, our $\delta$-hyperbolicity measurements are computed at the 
prompt level; corpus-level analysis may yield different insights.}
\section{Hyperbolic Fine-Tuning for LLMs}
\label{sec:HypLoRA_method}

\begin{table*}[t]
\small
\centering
\caption{Comparison of various LLMs on arithmetic reasoning tasks. The percentage following each dataset indicates the proportion of prompts relative to the total number of inference prompts. M.AVG represents the micro-average accuracy (since the datasets are imbalanced). For more adapter comparisons, please see Appendix~\ref{sec:full_comparison}.}
\resizebox{\textwidth}{!}{
\begin{tabular}{llcccccl}
\toprule
\textbf{Base Model} & \textbf{PEFT Method} & \textbf{\# Params (\%)} & \textbf{MAWPS(8.5\%)} & \textbf{SVAMP(35.6\%)} & \textbf{GSM8K(46.9\%)} & \textbf{AQuA(9.0\%)} & \textbf{M.AVG} \\ 
\midrule
GPT-3.5 & None & None & $87.4$ & $69.9$ & $56.4$ & $38.9$ & $62.3$ \\ 
\midrule
\multirow{2}{*}{LLaMA-7B} 
& LoRA & $0.83$ & $\textbf{81.9}$ & $48.2$ & $38.3$ & $18.5$ & $43.7$ \\
& \cellcolor{lightgray}\textbf{\method (Ours)} & \cellcolor{lightgray}$0.83$ & \cellcolor{lightgray}$79.0$ & \cellcolor{lightgray}$\textbf{49.1}$ & \cellcolor{lightgray}$\textbf{39.1}$ & \cellcolor{lightgray}$\textbf{20.5}$ & \cellcolor{lightgray}$\textbf{44.4}$ \\ 
\midrule
\multirow{2}{*}{LLaMA-13B} 
& LoRA & $0.67$ & $\textbf{83.5}$ & $54.7$ & $48.5$ & $18.5$ & $51.0$ \\
& \cellcolor{lightgray}\textbf{\method (Ours)} & \cellcolor{lightgray}$0.67$ & \cellcolor{lightgray}$83.2$ & \cellcolor{lightgray}$\textbf{54.8}$ & \cellcolor{lightgray}$\textbf{49.0}$ & \cellcolor{lightgray}$\textbf{21.5}$ & \cellcolor{lightgray}$\textbf{51.5}$ \\ 
\midrule
\multirow{2}{*}{Gemma-7B} 
& LoRA & $0.79$ & $\textbf{91.6}$ & $76.2$ & $66.3$ & $28.9$ & $68.6$ \\
& \cellcolor{lightgray}\textbf{\method (Ours)} & \cellcolor{lightgray}$0.79$ & \cellcolor{lightgray}${89.5}$ & \cellcolor{lightgray}$\textbf{78.7}$ & \cellcolor{lightgray}$\textbf{69.5}$ & \cellcolor{lightgray}$\textbf{32.7}$ & \cellcolor{lightgray}$\textbf{71.2}$ \\ 
\midrule
\multirow{2}{*}{LLaMA3-8B} 
& LoRA & $0.70$ & $\textbf{92.7}$ & ${78.9}$ & $70.8$ & $30.4$ & $71.9$ \\
& \cellcolor{lightgray}\textbf{\method (Ours)} & \cellcolor{lightgray}$0.70$ & 
\cellcolor{lightgray}$91.6$ & 
\cellcolor{lightgray}$\textbf{80.5}$ & \cellcolor{lightgray}$\textbf{74.0}$ & \cellcolor{lightgray}$\textbf{34.2}$ & \cellcolor{lightgray}$\textbf{74.2}$ \\ 
\midrule
\multirow{2}{*}{Gemma3-4B} 
& LoRA & $1.04$ & $\textbf{90.8}$ & $77.3$ & $72.3$ & $50.8$ & $73.7$ \\
& \cellcolor{lightgray}\textbf{\method (Ours)} & \cellcolor{lightgray}$1.04$ & \cellcolor{lightgray}$88.2$ & \cellcolor{lightgray}$\textbf{83.9}$ & \cellcolor{lightgray}$\textbf{76.1}$ & \cellcolor{lightgray}$\textbf{53.2}$ & \cellcolor{lightgray}$\textbf{77.8}$ \\ 
\midrule
\multirow{2}{*}{Qwen2.5-7B} 
& LoRA & $0.71$ & ${90.8}$ & $84.4$ & $78.6$ & ${68.1}$ & $80.8$ \\
& \cellcolor{lightgray}\textbf{\method (Ours)} & \cellcolor{lightgray}$0.71$ & \cellcolor{lightgray}$\textbf{91.2}$ & \cellcolor{lightgray}$\textbf{92.2}$ & \cellcolor{lightgray}$\textbf{87.9}$ & \cellcolor{lightgray}$\textbf{71.6}$ & \cellcolor{lightgray}$\textbf{88.3}$ \\ 
\bottomrule
\end{tabular}
}
\label{tab:main_results}
\vspace{-10pt}
\end{table*}

The core technique in the LoRA adapter involves matrix transformations. The conventional approach to implementing these transformations in the Lorentz model of hyperbolic geometry is through operations in the tangent space, while maintaining the learnable weights in Euclidean space~\citep{ganea2018hyperbolic_hnn,chami2019hyperbolic}. However, this approach presents a significant challenge for our application. Since the hidden states of LLMs exist in Euclidean space, we would need to project these states to hyperbolic space and subsequently map them back to the tangent space. This process results in consecutive logarithmic and exponential mappings $(\log_\mathbf{o}^K(\exp_\mathbf{o}^K(\mathbf{x})))$, which effectively cancel each other out, reducing the method to the original LoRA approach and nullifying any benefits from hyperbolic geometry.

\textbf{Direct Lorentz Low-Rank Transformation (LLR)}. To overcome this limitation, we propose a direct Lorentz Low-Rank Transformation (LLR) that operates directly on the hyperbolic space without relying on tangent space mappings. This approach allows us to perform low-rank adaptation while preserving the advantages of hyperbolic geometry:
\begin{equation}
\begin{aligned}
\mathbf{z}^E &= W_\text{LoRA}(\mathbf{x}^E) = W\mathbf{x}^E + \Delta W\mathbf{x}^E \\
&= W\mathbf{x}^E + \Pi_{\log}^K(\mathbf{LLR}(BA, \Pi_{\exp}^K(\mathbf{x}^E))),
\end{aligned}
\end{equation}
where $\mathbf{LLR}$ represents the direct Lorentz Low-Rank Transformation that operates directly on the hyperbolic representation $\mathbf{x}^H = \Pi_{\exp}^K(\mathbf{x}^E)$:

\begin{equation}
\mathbf{LLR}(BA, \mathbf{x}^H) = (\sqrt{\|{BA\mathbf{x}_s^H}\|^2_2 + K}, {BA\mathbf{x}_s^H}), 
\end{equation}
where $\mathbf{x}_s^H$ is the space-like component of $\mathbf{x}^H$, i.e., $\mathbf{x}_s^H=\mathbf{x}_{[1:n]}^H$ without the first time-like dimension $\mathbf{x}_{[0:1]}^H$. The operators $\Pi_{\exp}^K$ and $\Pi_{\log}^K$ represent projections from Euclidean space to hyperbolic space and from hyperbolic space to Euclidean space, respectively. 
The detailed formulas are provided in Appendix~\ref{appendix:exponential_and_logarithmic_map}. 
It can be verified that $\mathbf{LLR}(BA, \mathbf{x}^H) \in \mathcal{L}^n$, ensuring that our transformation remains within the Lorentz model of hyperbolic space. This transformation primarily affects the space-like dimensions, functioning similarly to a pseudo-Lorentz rotation~\cite{chen2021fully}.
The linear transformation is inspired by hyperbolic neural networks~\citep{chen2021fully,yang2024hypformer,dai2021hyperbolic}.
For efficient integration with LLMs, the transformation removes normalization and non-linear activation terms in \citep{chen2021fully}, varying curvatures in \cite{yang2024hypformer}, and orthogonal constraints in \citep{dai2021hyperbolic}. Our main contribution lies in applying hyperbolic low-rank adaptation for LLMs, while the specific linear transformation itself is flexible, and other transformations on the manifold could also be compatible with our approach.

By adapting in the hyperbolic domain, \method captures more complex hierarchical relationships than traditional Euclidean-based methods, as detailed in Proposition~\ref{prop:hyp_effectiveness}. Additionally, the low-rank nature of the adaptation matrices ${A}$ and ${B}$ promotes parameter efficiency, making \method well-suited for LLMs.

\begin{table}[!t]
    \centering
    \caption{Comparison of various LLMs on commonsense reasoning tasks. These datasets contain relatively similar amounts of data, so we use AVG to represent the average accuracy.}
    \vspace{2pt}
    \resizebox{\textwidth}{!}{
    \begin{tabular}{@{}llcccccccccc@{}}
    \toprule
    \textbf{Base Model} & \textbf{PEFT Method} & \textbf{\# Params (\%)} & \textbf{BoolQ} & \textbf{PIQA} & \textbf{SIQA} & \textbf{HellaSwag} & \textbf{WinoGrande} & \textbf{ARC-e} & \textbf{ARC-c} & \textbf{OBQA} & \textbf{AVG} \\
    \midrule
    GPT-3.5 & None & None & $73.1$ & $85.4$ & $68.5$ & $78.5$ & $66.1$ & $89.8$ & $79.9$ & $74.8$ & $77.0$ \\
    \midrule
    \multirow{2}{*}{LLaMA3-8B} & LoRA & $0.70$ & $70.8$ & $85.2$ & $79.9$ & $91.7$ & $84.3$ & $84.2$ & $71.2$ & $79.0$ & $80.8$ \\
    & \cellcolor{lightgray}\textbf{HypLoRA (Ours)}
    & \cellcolor{lightgray}$0.70$
    & \cellcolor{lightgray}$\textbf{74.1}$
    & \cellcolor{lightgray}$\textbf{87.6}$
    & \cellcolor{lightgray}$\textbf{80.6}$
    & \cellcolor{lightgray}$\textbf{94.5}$
    & \cellcolor{lightgray}$\textbf{84.7}$
    & \cellcolor{lightgray}$\textbf{90.4}$
    & \cellcolor{lightgray}$\textbf{81.2}$
    & \cellcolor{lightgray}$\textbf{85.2}$
    & \cellcolor{lightgray}$\textbf{84.8}$ \\
    \midrule
    \multirow{2}{*}{Gemma3-4B} & LoRA & $1.04$ & $68.1$ & $83.2$ & $77.2$ & $88.9$ & $\textbf{80.5}$ & $84.5$ & $69.9$ & $83.6$ & $79.5$ \\
    & \cellcolor{lightgray}\textbf{HypLoRA (Ours)}
    & \cellcolor{lightgray}$1.04$
    & \cellcolor{lightgray}$\textbf{70.0}$
    & \cellcolor{lightgray}$\textbf{84.3}$
    & \cellcolor{lightgray}$\textbf{79.2}$
    & \cellcolor{lightgray}$\textbf{91.5}$
    & \cellcolor{lightgray}$80.3$
    & \cellcolor{lightgray}$\textbf{89.1}$
    & \cellcolor{lightgray}$\textbf{75.9}$
    & \cellcolor{lightgray}$\textbf{86.4}$
    & \cellcolor{lightgray}$\textbf{82.5}$ \\
    \midrule
    \multirow{2}{*}{Qwen2.5-7B} & LoRA & $0.71$ & $\textbf{73.4}$ & $\textbf{89.5}$ & $79.5$ & $93.6$ & $84.1$ & $92.8$ & $82.0$ & $87.0$ & $85.2$ \\
    & \cellcolor{lightgray}\textbf{HypLoRA (Ours)}
    & \cellcolor{lightgray}$0.71$
    & \cellcolor{lightgray}$72.8$
    & \cellcolor{lightgray}$89.3$
    & \cellcolor{lightgray}$\textbf{79.8}$
    & \cellcolor{lightgray}$\textbf{94.8}$
    & \cellcolor{lightgray}$\textbf{84.4}$
    & \cellcolor{lightgray}$\textbf{95.5}$
    & \cellcolor{lightgray}$\textbf{87.5}$
    & \cellcolor{lightgray}$\textbf{90.8}$
    & \cellcolor{lightgray}$\textbf{87.0}$ \\
    \bottomrule
    \end{tabular}
    }
    \label{tab:peft_comparison_common}
    \vspace{-10pt}
    \end{table}

\textbf{Time Complexity Analysis.} \method has similar theoretical time complexity as the Euclidean LoRA, which is $\mathcal{O}(r \cdot (d+k))$, where $d$ and $k$ represent the input and output dimensions, respectively. However, in practical implementation, \method introduces additional computations due to the space mapping. These additional operations, nevertheless, can be completed within $\mathcal{O}(N)$ where $N$ is the number of input tokens.

\begin{proposition}
\label{prop:hyp_effectiveness}
Let $\mathbf{x} \in \mathbb{R}^d$ denote the input token embeddings. The HypLoRA adaptation, applied to $\mathbf{x}$, involves a sequence of projection into hyperbolic space, a Direct Lorentz Low-Rank Transformation (LLR), and projection back to Euclidean space. Due to the non-linear nature of these hyperbolic operations, the effective transformation applied by HypLoRA introduces higher-order terms with respect to $\mathbf{x}$. As detailed in Appendix~\ref{appendix:theoretical_analysis}, these terms exhibit explicit dependency on the L2 norm, $\|\mathbf{x}\|_2$, of the input embeddings. This norm-dependent, higher-order modification enables HypLoRA to capture hierarchical relationships in the embedding space, thereby achieving natural alignment with the underlying hyperbolic geometry of the token representations.

\end{proposition}

\subsection{Experimental Settings}
\label{equ:experiments_settings}

\textbf{Datasets.} Following the experiment setup outlined in~\citep{hu2023llm}, we utilize two high-quality datasets, Math10K and Commonsense170K, tailored for mathematical and commonsense reasoning, respectively. Math10K consists of training data from GSM8K~\citep{cobbe2021training}, MAWPS, MAWPS-single~\citep{koncel2016MAWPS}, and 1,000 samples from AQuA~\citep{ling2017program}, augmented with ChatGPT-generated step-by-step rationales to reinforce reasoning capabilities. The test set includes GSM8K, AQuA, MAWPS, and SVAMP~\citep{patel2021nlp}, ensuring no overlap with the training data. Commonsense170K is constructed by reformatting samples from BoolQ, PIQA, SIQA, HellaSwag, WinoGrande, ARC-e, ARC-c, and OBQA using standardized templates that outline the task, content, and answer, resulting in 170K training samples. The test datasets are drawn from the same sources, with strict separation from training samples. 
For fine-tuning methods, we compare with LoRA~\citep{hu2021lora} and also make a comparison with other adapters in Appendix~\ref{sec:full_comparison}, which also includes training details.

\subsection{Experimental Results}

Table~\ref{tab:main_results} summarizes our key experimental outcomes on arithmetic reasoning tasks, while Table~\ref{tab:peft_comparison_common} presents results for commonsense reasoning benchmarks. Our primary comparison contrasts LoRA and \method to demonstrate the effectiveness of the proposed approach, with additional baselines provided in Appendix~\ref{sec:full_comparison}.

\textbf{Arithmetic Reasoning Performance.} On arithmetic reasoning tasks, as indicated by results in Table~\ref{tab:main_results}, \method shows notable efficacy, especially on datasets recognized for their complexity, such as GSM8K, AQuA, and SVAMP. These datasets demand robust multi-step reasoning and a nuanced understanding of numerical and textual relationships. For instance, on Qwen2.5-7B, \method achieves a substantial $+7.5$ percentage point improvement in M.AVG ($88.3\%$ vs. $80.8\%$), with notable gains of $+7.8\%$ on SVAMP and $+9.3\%$ on GSM8K. 
The enhanced performance of \method in these areas aligns with its design; by operating in hyperbolic space, it can better model the hierarchical structure of problems and distinguish subtle yet critical differences in input embeddings. This is further corroborated by the theoretical analysis (Appendix~\ref{appendix:theoretical_analysis}), which posits that \method introduces higher-order, norm-dependent terms. These terms allow the model to develop a more refined sensitivity to token importance and inter-token relationships.

\textbf{Commonsense Reasoning Performance.} The robust performance of \method extends to commonsense reasoning, as detailed in Table~\ref{tab:peft_comparison_common}. For the Gemma3-4B model, \method achieved an average accuracy of 82.5\% across all datasets, surpassing LoRA's 79.5\%. Similarly, on the Qwen2.5-7B model, \method obtained an average of 87.0\% compared to LoRA's 85.2\%. These improvements are distributed across various commonsense benchmarks, including notable gains on datasets like ARC-c and OBQA for Gemma3-4B, and ARC-c, ARC-e, and OBQA for Qwen2.5-7B. Commonsense reasoning often relies on understanding implicit relationships and contextual nuances, which may not always be explicitly hierarchical but still benefit from the richer representational capacity offered by hyperbolic geometry. The ability of \method to better discern these subtleties, likely due to the mechanisms described in Proposition~\ref{prop:hyp_effectiveness}, contributes to these observed performance gains, showcasing the broad applicability of hyperbolic fine-tuning.

\begin{figure*}[t]
    \centering
    \begin{minipage}[t]{0.32\textwidth} 
        \centering
        \captionof{table}{Results for varying curvature $K$ on the Gemma3-4B model}
        \label{tab:curvature_comparsions} 
        \resizebox{0.8\textwidth}{!}{
        \begin{tabular}{lccc}
        \toprule
        {Dataset} & {K=$0.5$} & {K=$1.0$}\\
        \midrule
        MAWPS & $88.2$ & $\textbf{91.9}$\\
        SVAMP & $\textbf{83.9}$ & $80.3$\\
        GSM8K & $\textbf{76.1}$ & $73.8$\\
        AQuA  & $\textbf{53.5}$ & $52.7$\\
        \midrule
        M.AVG & $\textbf{77.8}$ & $75.8$ \\
        \bottomrule
        \end{tabular}
        }
    \end{minipage}%
    \hspace{20pt}
    \begin{minipage}[t]{0.6\textwidth} 
        \centering
        \caption{GPU (A100) usage during inference}
        \label{fig:gpu_usage}
        \includegraphics[width=0.48\linewidth]{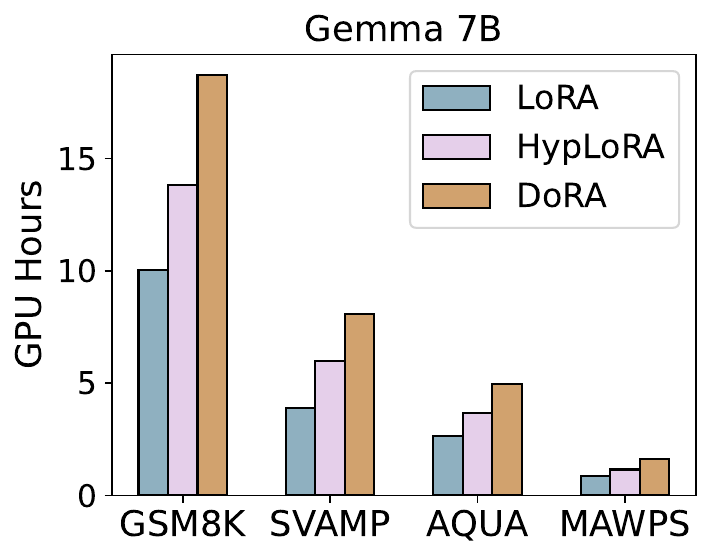}
        \hfill
         \includegraphics[width=0.48\linewidth]{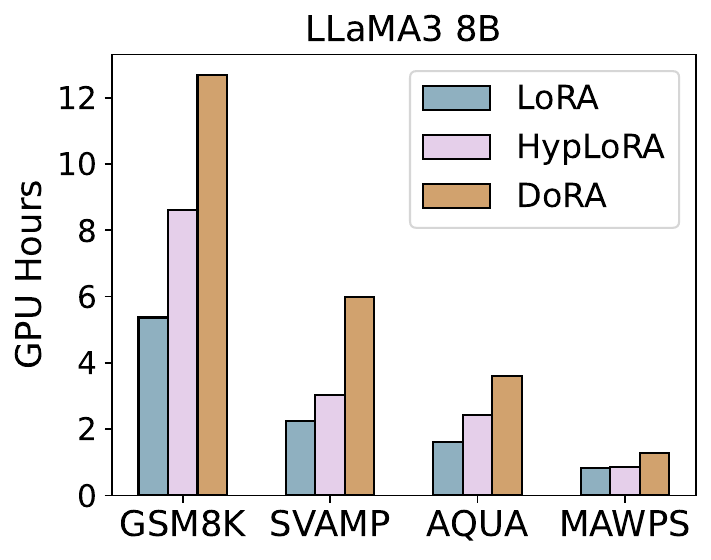}
    \end{minipage}
    \vspace{-15pt}
\end{figure*}

\textbf{The Impact of Curvature on Performance.}
Curvature in hyperbolic space is a key hyperparameter in \method, directly affecting its capacity to model underlying structures and geometries. To evaluate its impact, we experiment with a learnable curvature initialized with different curvature values on the Gemma3-4B model, as shown in Table~\ref{tab:curvature_comparsions}, where the curvature is defined as ${-1}/{K}$.
Our results demonstrate that curvature does influence model performance. For Gemma-7B and Gemma3-4B, a curvature value of $0.5$ consistently yields the best overall performance across both arithmetic and commonsense reasoning benchmarks. Similarly, for LLaMA3-8B, $0.5$ proves optimal. In commonsense reasoning benchmarks, a curvature of $1.0$ performs best for LLaMA3-8B and Qwen2.5-7B.

\textbf{Efficiency.} In Section~\ref{sec:HypLoRA_method}, we analyze the time complexity of our approach, which remains consistent with that of LoRA. However, during actual inference, \method incurs additional computational overhead due to operations such as projections. These operations introduce some additional runtime, particularly for larger models. 
The GPU hours for inference on four datasets are presented in Figure~\ref{fig:gpu_usage}.
Despite this overhead, our method demonstrates improved efficiency when compared to the previous competitive model, DoRA. Notably, \method still outperforms DoRA in terms of both runtime and overall efficiency. Besides, all models can be fine-tuned in approximately one hour for optimal training efficiency.

\section{Conclusion}
In this work, we investigated the non-Euclidean geometric properties inherent in LLMs, confirming their strong hyperbolic characteristics, which suggest underlying hierarchical structures. Building on these insights, we introduced \method, a hyperbolic low-rank adaptation technique. \method performs fine-tuning directly on the hyperbolic manifold.
Extensive experiments show that \method significantly improves LLM performance on arithmetic reasoning and commonsense tasks. By leveraging the hyperbolic structure of the data, \method enhances the model's ability to capture and utilize intricate relationships, leading to better reasoning capabilities.

\textbf{Broader Impact.} Enhancing reasoning-oriented LLMs can help education, scientific assistance, and safer decision-support systems, but the same improvements may also accelerate misuse (e.g., automating complex disinformation or amplifying biased advice) and increase energy consumption due to added hyperbolic projections. We therefore advocate releasing checkpoints and code with usage guidelines (as in our public repo), tracking compute budgets when scaling HypLoRA further.

\section*{Acknowledgment}
The authors would like to express their sincere gratitude to the anonymous reviewers and the Area Chair for their valuable comments and insightful suggestions, which have greatly improved this work. 
We thank our colleagues and lab mates at HKUST(GZ), IISc, Yale, Stanford, and CUHK for motivating discussions on geometric learning for LLMs, and we acknowledge the administrators of the institutional GPU clusters that enabled the large-scale experiments. Part of this research was conducted while Menglin Yang was at Yale University, whose support and computing resources were instrumental in shaping the initial ideas of this project.

\newpage
\bibliographystyle{unsrt} 
\bibliography{references}

@article{achiam2023gpt,
 author = {Achiam, Josh and Adler, Steven and Agarwal, Sandhini and Ahmad, Lama and Akkaya, Ilge and Aleman, Florencia Leoni and Almeida, Diogo and Altenschmidt, Janko and Altman, Sam and Anadkat, Shyamal and others},
 journal = {arXiv preprint arXiv:2303.08774},
 title = {{GPT-4 Technical Report}},
 year = {2023}
}

@article{mao2024klein,
  title={Klein Model for Hyperbolic Neural Networks},
  author={Mao, Yidan and Gu, Jing and Werner, Marcus C and Zou, Dongmian},
  journal={arXiv preprint arXiv:2410.16813},
  year={2024}
}

@inproceedings{gao2021curvature,
  title={Curvature generation in curved spaces for few-shot learning},
  author={Gao, Zhi and Wu, Yuwei and Jia, Yunde and Harandi, Mehrtash},
  booktitle={IEEE/CVF International Conference on Computer Vision (ICCV)},
  pages={8691--8700},
  year={2021}
}

@article{fan2025curvature,
  title={Curvature Learning for Generalization of Hyperbolic Neural Networks},
  author={Fan, Xiaomeng and Wu, Yuwei and Gao, Zhi and Harandi, Mehrtash and Jia, Yunde},
  journal={International Journal of Computer Vision (IJCV)},
  pages={1--37},
  year={2025}
}

@article{aghajanyan2020intrinsic,
 author = {Aghajanyan, Armen and Zettlemoyer, Luke and Gupta, Sonal},
 journal = {arXiv preprint arXiv:2012.13255},
 title = {Intrinsic dimensionality explains the effectiveness of language model fine-tuning},
 year = {2020}
}

@article{alstott2014powerlaw,
 author = {Alstott, Jeff and Bullmore, Ed and Plenz, Dietmar},
 journal = {PloS one},
 number = {1},
 pages = {e85777},
 title = {Powerlaw: a Python package for analysis of heavy-tailed distributions},
 volume = {9},
 year = {2014}
}

@article{alvarez2006hierarchical,
  author    = {Alvarez-Lacalle, Enrique and Dorow, Beate and Eckmann, J-P and Moses, Elisha},
  title     = {Hierarchical structures induce long-range dynamical correlations in written texts},
  journal   = {Proceedings of the National Academy of Sciences},
  volume    = {103},
  number    = {21},
  pages     = {7956--7961},
  year      = {2006}
}

@inproceedings{bachmann2020constant,
 author = {Bachmann, Gregor and B{\'e}cigneul, Gary and Ganea, Octavian},
 booktitle = {International Conference on Machine Learning (ICML)},
 organization = {PMLR},
 pages = {486--496},
 title = {Constant curvature graph convolutional networks},
 year = {2020}
}

@inproceedings{barabasi2003scale,
 author = {Barab{\'a}si, Albert-L{\'a}szl{\'o} and Dezs{\H{o}}, Zolt{\'a}n and Ravasz, Erzs{\'e}bet and Yook, Soon-Hyung and Oltvai, Zolt{\'a}n},
 booktitle = {AIP Conference Proceedings},
 number = {1},
 organization = {American Institute of Physics},
 pages = {1--16},
 title = {Scale-free and hierarchical structures in complex networks},
 volume = {661},
 year = {2003}
}

@article{bdeir2023hyperbolic,
 author = {Bdeir, Ahmad and Schwethelm, Kristian and Landwehr, Niels},
 journal = {arXiv preprint arXiv:2303.15919},
 title = {Hyperbolic geometry in computer vision: A novel framework for convolutional neural networks},
 year = {2023}
}

@article{becigneul2018riemannian,
 author = {B{\'e}cigneul, Gary and Ganea, Octavian-Eugen},
 journal = {arXiv preprint arXiv:1810.00760},
 title = {Riemannian adaptive optimization methods},
 year = {2018}
}

@article{borassi2015hyperbolicity,
 author = {Borassi, Michele and Chessa, Alessandro and Caldarelli, Guido},
 journal = {Physical Review E},
 number = {3},
 pages = {032812},
 title = {Hyperbolicity measures democracy in real-world networks},
 volume = {92},
 year = {2015}
}

@article{bronstein2017geometric,
 author = {Bronstein, Michael M and Bruna, Joan and LeCun, Yann and Szlam, Arthur and Vandergheynst, Pierre},
 journal = {IEEE Signal Processing Magazine},
 number = {4},
 pages = {18--42},
 title = {Geometric deep learning: going beyond euclidean data},
 volume = {34},
 year = {2017}
}

@article{cannon1997hyperbolic,
 author = {Cannon, James W and Floyd, William J and Kenyon, Richard and Parry, Walter R and others},
 journal = {Flavors of geometry},
 number = {59-115},
 pages = {2},
 title = {Hyperbolic geometry},
 volume = {31},
 year = {1997}
}

@article{cetin2022hyperbolic,
 author = {Cetin, Edoardo and Chamberlain, Benjamin and Bronstein, Michael and Hunt, Jonathan J},
 journal = {arXiv preprint arXiv:2210.01542},
 title = {Hyperbolic deep reinforcement learning},
 year = {2022}
}

@inproceedings{chami2019hyperbolic,
 author = {Chami, Ines and Ying, Zhitao and R{\'e}, Christopher and Leskovec, Jure},
 booktitle = {Conference on Neural Information Processing Systems (NeurIPS)},
 title = {Hyperbolic graph convolutional neural networks},
 volume = {32},
 year = {2019}
}

@article{chen2021fully,
 author = {Chen, Weize and Han, Xu and Lin, Yankai and Zhao, Hexu and Liu, Zhiyuan and Li, Peng and Sun, Maosong and Zhou, Jie},
 journal = {arXiv preprint arXiv:2105.14686},
 title = {Fully Hyperbolic Neural Networks},
 year = {2021}
}

@article{cho2023curve,
 author = {Cho, Sungjun and Cho, Seunghyuk and Park, Sungwoo and Lee, Hankook and Lee, Honglak and Lee, Moontae},
 journal = {arXiv preprint arXiv:2309.04082},
 title = {Curve your attention: Mixed-curvature transformers for graph representation learning},
 year = {2023}
}

@article{cobbe2021training,
 author = {Cobbe, Karl and Kosaraju, Vineet and Bavarian, Mohammad and Chen, Mark and Jun, Heewoo and Kaiser, Lukasz and Plappert, Matthias and Tworek, Jerry and Hilton, Jacob and Nakano, Reiichiro and others},
 journal = {arXiv preprint arXiv:2110.14168},
 title = {Training verifiers to solve math word problems},
 year = {2021}
}

@inproceedings{dai2021hyperbolic,
 author = {Dai, Jindou and Wu, Yuwei and Gao, Zhi and Jia, Yunde},
 booktitle = {IEEE/CVF Conference on Computer Vision and Pattern Recognition (CVPR)},
 pages = {154--163},
 title = {A hyperbolic-to-hyperbolic graph convolutional network},
 year = {2021}
}

@inproceedings{desai2023meru,
 author = {Desai, Karan and Nickel, Maximilian and Rajpurohit, Tanmay and Johnson, Justin and Vedantam, Ramakrishna},
 booktitle = {International Conference on Machine Learning (ICML)},
 title = {{Hyperbolic image-text representations}},
 year = {2023}
}

@inproceedings{dettmers2024qlora,
 author = {Dettmers, Tim and Pagnoni, Artidoro and Holtzman, Ari and Zettlemoyer, Luke},
 booktitle = {Conference on Neural Information Processing Systems (NeurIPS)},
 title = {Qlora: Efficient finetuning of quantized llms},
 volume = {36},
 year = {2024}
}

@article{edalati2022krona,
 author = {Edalati, Ali and Tahaei, Marzieh and Kobyzev, Ivan and Nia, Vahid Partovi and Clark, James J and Rezagholizadeh, Mehdi},
 journal = {arXiv preprint arXiv:2212.10650},
 title = {Krona: Parameter efficient tuning with kronecker adapter},
 year = {2022}
}

@article{fournier2015computing,
 author = {Fournier, Herv{\'e} and Ismail, Anas and Vigneron, Antoine},
 journal = {Information Processing Letters},
 number = {6-8},
 pages = {576--579},
 title = {Computing the Gromov hyperbolicity of a discrete metric space},
 volume = {115},
 year = {2015}
}

@article{fu2024hyperbolic,
 author = {Fu, Xingcheng and Gao, Yisen and Wei, Yuecen and Sun, Qingyun and Peng, Hao and Li, Jianxin and Li, Xianxian},
 journal = {arXiv preprint arXiv:2405.03188},
 title = {Hyperbolic geometric latent diffusion model for graph generation},
 year = {2024}
}

@inproceedings{ganea2018hyperbolic,
 author = {Ganea, Octavian and B{\'e}cigneul, Gary and Hofmann, Thomas},
 booktitle = {International Conference on Machine Learning (ICML)},
 organization = {PMLR},
 pages = {1646--1655},
 title = {Hyperbolic entailment cones for learning hierarchical embeddings},
 year = {2018}
}

@inproceedings{ganea2018hyperbolic_hnn,
 author = {Ganea, Octavian and B{\'e}cigneul, Gary and Hofmann, Thomas},
 booktitle = {Conference on Neural Information Processing Systems (NeurIPS)},
 title = {Hyperbolic neural networks},
 volume = {31},
 year = {2018}
}

@article{gao2019representation,
 author = {Gao, Jun and He, Di and Tan, Xu and Qin, Tao and Wang, Liwei and Liu, Tie-Yan},
 journal = {arXiv preprint arXiv:1907.12009},
 title = {Representation degeneration problem in training natural language generation models},
 year = {2019}
}

@article{google2024gemma,
 author = {Gemma Team, Google Deepmind},
 journal = {arXiv preprint arXiv:2403.08295},
 title = {Gemma: Open models based on Gemini research and technology},
 year = {2024}
}

@incollection{gromov1987hyperbolic,
 author = {Gromov, Mikhael},
 booktitle = {Essays in group theory},
 pages = {75--263},
 publisher = {Springer},
 title = {Hyperbolic groups},
 year = {1987}
}

@article{gulcehre2018hyperbolic,
 author = {Gulcehre, Caglar and Denil, Misha and Malinowski, Mateusz and Razavi, Ali and Pascanu, Razvan and Hermann, Karl Moritz and Battaglia, Peter and Bapst, Victor and Raposo, David and Santoro, Adam and others},
 journal = {arXiv preprint arXiv:1805.09786},
 title = {Hyperbolic attention networks},
 year = {2018}
}

@inproceedings{hagberg2008exploring,
 author = {Hagberg, Aric A. and Schult, Daniel A. and Swart, Pieter J.},
 booktitle = {SciPy2008},
 editor = {Varoquaux, Gäel and Vaught, Travis and Millman, Jarrod},
 organization = {Pasadena, CA USA},
 pages = {11--15},
 title = {Exploring network structure, dynamics, and function using NetworkX},
 year = {2008}
}

@article{hayou2024lora+,
 author = {Hayou, Soufiane and Ghosh, Nikhil and Yu, Bin},
 journal = {arXiv preprint arXiv:2402.12354},
 title = {LoRA+: Efficient low rank adaptation of large models},
 year = {2024}
}

@inproceedings{he2025hypercore,
 author = {Neil He and Menglin Yang and Rex Ying},
 booktitle = {International Conference on Learning Representations (ICLR)},
 title = {HyperCore: The core framework for building hyperbolic foundation models with comprehensive modules},
 url = {https://arxiv.org/abs/2504.08912},
 year = {2025}
}

@inproceedings{houlsby2019parameter,
 author = {Houlsby, Neil and Giurgiu, Andrei and Jastrzebski, Stanislaw and Morrone, Bruna and De Laroussilhe, Quentin and Gesmundo, Andrea and Attariyan, Mona and Gelly, Sylvain},
 booktitle = {International Conference on Machine Learning (ICML)},
 organization = {PMLR},
 pages = {2790--2799},
 title = {Parameter-efficient transfer learning for NLP},
 year = {2019}
}

@article{hu2021lora,
 author = {Hu, Edward J and Shen, Yelong and Wallis, Phillip and Allen-Zhu, Zeyuan and Li, Yuanzhi and Wang, Shean and Wang, Lu and Chen, Weizhu},
 journal = {arXiv preprint arXiv:2106.09685},
 title = {Lora: Low-rank adaptation of large language models},
 year = {2021}
}

@article{hu2023llm,
 author = {Hu, Zhiqiang and Lan, Yihuai and Wang, Lei and Xu, Wanyu and Lim, Ee-Peng and Lee, Roy Ka-Wei and Bing, Lidong and Poria, Soujanya},
 journal = {arXiv preprint arXiv:2304.01933},
 title = {LLM-Adapters: An adapter family for parameter-efficient fine-Tuning of large language models},
 year = {2023}
}

@article{kennedy2013hyperbolicity,
 author = {Kennedy, W Sean and Narayan, Onuttom and Saniee, Iraj},
 journal = {arXiv preprint arXiv:1307.0031},
 title = {On the hyperbolicity of large-scale networks},
 year = {2013}
}

@article{khan2025hyperbolic,
 author = {Khan, Raiyan R and Chlenski, Philippe and Pe'er, Itsik},
 journal = {arXiv preprint arXiv:2507.21648},
 title = {Hyperbolic genome embeddings},
 year = {2025}
}

@inproceedings{khrulkov2020hyperbolic,
 author = {Khrulkov, Valentin and Mirvakhabova, Leyla and Ustinova, Evgeniya and Oseledets, Ivan and Lempitsky, Victor},
 booktitle = {IEEE/CVF Conference on Computer Vision and Pattern Recognition (CVPR)},
 pages = {6418--6428},
 title = {Hyperbolic image embeddings},
 year = {2020}
}

@article{kochurov2020geoopt,
 author = {Kochurov, Max and Karimov, Rasul and Kozlukov, Serge},
 journal = {arXiv preprint arXiv:2005.02819},
 title = {Geoopt: Riemannian optimization in pytorch},
 year = {2020}
}

@inproceedings{koncel2016mawps,
 author = {Koncel-Kedziorski, Rik and Roy, Subhro and Amini, Aida and Kushman, Nate and Hajishirzi, Hannaneh},
 booktitle = {Annual Conference of the North American Chapter of the Association for Computational Linguistics (NAACL)},
 pages = {1152--1157},
 title = {MAWPS: A math word problem repository},
 year = {2016}
}

@article{krioukov2009curvature,
 author = {Krioukov, Dmitri and Papadopoulos, Fragkiskos and Vahdat, Amin and Bogun{\'a}, Mari{\'a}n},
 journal = {Physical Review E},
 number = {3},
 pages = {035101},
 title = {Curvature and temperature of complex networks},
 volume = {80},
 year = {2009}
}

@article{krioukov2010hyperbolic,
 author = {Krioukov, Dmitri and Papadopoulos, Fragkiskos and Kitsak, Maksim and Vahdat, Amin and Bogun{\'a}, Mari{\'a}n},
 journal = {Physical Review E—Statistical, Nonlinear, and Soft Matter Physics},
 number = {3},
 pages = {036106},
 title = {Hyperbolic geometry of complex networks},
 volume = {82},
 year = {2010}
}

@inproceedings{yang2022hrcf,
  title={HRCF: Enhancing collaborative filtering via hyperbolic geometric regularization},
  author={Yang, Menglin and Zhou, Min and Liu, Jiahong and Lian, Defu and King, Irwin},
  booktitle={The Web Conference (WWW)},
  pages={2462--2471},
  year={2022}
}

@article{zhang2021hyperbolic,
  title={Hyperbolic graph attention network},
  author={Zhang, Yiding and Wang, Xiao and Shi, Chuan and Jiang, Xunqiang and Ye, Yanfang},
  journal={IEEE Transactions on Big Data},
  volume={8},
  number={6},
  pages={1690--1701},
  year={2021},
  publisher={IEEE}
}

@article{lester2021power,
 author = {Lester, Brian and Al-Rfou, Rami and Constant, Noah},
 journal = {arXiv preprint arXiv:2104.08691},
 title = {The power of scale for parameter-efficient prompt tuning},
 year = {2021}
}

@article{li2021prefix,
 author = {Li, Xiang Lisa and Liang, Percy},
 journal = {arXiv preprint arXiv:2101.00190},
 title = {Prefix-tuning: Optimizing continuous prompts for generation},
 year = {2021}
}

@article{li2023loftq,
 author = {Li, Yixiao and Yu, Yifan and Liang, Chen and He, Pengcheng and Karampatziakis, Nikos and Chen, Weizhu and Zhao, Tuo},
 journal = {arXiv preprint arXiv:2310.08659},
 title = {Loftq: Lora-fine-tuning-aware quantization for large language models},
 year = {2023}
}

@article{ling2017program,
 author = {Ling, Wang and Yogatama, Dani and Dyer, Chris and Blunsom, Phil},
 journal = {arXiv preprint arXiv:1705.04146},
 title = {Program induction by rationale generation: Learning to solve and explain algebraic word problems},
 year = {2017}
}

@inproceedings{liu2019hyperbolic,
 author = {Liu, Qi and Nickel, Maximilian and Kiela, Douwe},
 booktitle = {Conference on Neural Information Processing Systems (NeurIPS)},
 title = {Hyperbolic graph neural networks},
 volume = {32},
 year = {2019}
}

@inproceedings{liu2024client,
 author = {Liu, Jiahong and Fu, Xinyu and Yang, Menglin and Zhang, Weixi and Ying, Rex and King, Irwin},
 booktitle = {ACM SIGKDD Conference on Knowledge Discovery and Data Mining (KDD)},
 title = {Client-specific hyperbolic federated learning},
 year = {2024}
}

@article{liu2024dora,
 author = {Liu, Shih-Yang and Wang, Chien-Yi and Yin, Hongxu and Molchanov, Pavlo and Wang, Yu-Chiang Frank and Cheng, Kwang-Ting and Chen, Min-Hung},
 journal = {arXiv preprint arXiv:2402.09353},
 title = {DoRA: Weight-decomposed low-rank adaptation},
 year = {2024}
}

@article{liu2025survey,
 author = {Liu, Jiahong and Qiu, Zexuan and Li, Zhongyang and Dai, Quanyu and Yu, Wenhao and Zhu, Jieming and Hu, Minda and Yang, Menglin and Chua, Tat-Seng and King, Irwin},
 journal = {arXiv preprint arXiv:2502.11528},
 title = {A survey of personalized large language models: Progress and future directions},
 year = {2025}
}

@article{loshchilov2017decoupled,
 author = {Loshchilov, Ilya and Hutter, Frank},
 journal = {arXiv preprint arXiv:1711.05101},
 title = {Decoupled weight decay regularization},
 year = {2017}
}

@article{mettes2023hyperbolic,
 author = {Mettes, Pascal and Atigh, Mina Ghadimi and Keller-Ressel, Martin and Gu, Jeffrey and Yeung, Serena},
 journal = {arXiv preprint arXiv:2305.06611},
 title = {Hyperbolic deep learning in computer vision: A Survey},
 year = {2023}
}

@article{mirzadeh2024gsm,
 author = {Mirzadeh, Iman and Alizadeh, Keivan and Shahrokhi, Hooman and Tuzel, Oncel and Bengio, Samy and Farajtabar, Mehrdad},
 journal = {arXiv preprint arXiv:2410.05229},
 title = {Gsm-symbolic: Understanding the limitations of mathematical reasoning in large language models},
 year = {2024}
}

@inproceedings{mishne2023numerical,
 author = {Mishne, Gal and Wan, Zhengchao and Wang, Yusu and Yang, Sheng},
 booktitle = {International Conference on Machine Learning (ICML)},
 organization = {PMLR},
 pages = {24925--24949},
 title = {The numerical stability of hyperbolic representation learning},
 year = {2023}
}

@article{moretti2002interplay,
 author = {Moretti, Valter},
 journal = {arXiv preprint math-ph/0211047},
 title = {The interplay of the polar decomposition theorem and the Lorentz group},
 year = {2002}
}

@book{morse1946methods,
 author = {Morse, Philip McCord and Feshbach, Herman},
 publisher = {Technology Press},
 title = {Methods of theoretical physics},
 year = {1946}
}

@article{nakaishi2025rethinking,
 author = {Nakaishi, Kai and Yoshida, Ryo and Kajikawa, Kohei and Hukushima, Koji and Oseki, Yohei},
 journal = {arXiv preprint arXiv:2505.04984},
 title = {Rethinking the relationship between the power law and hierarchical structures},
 year = {2025}
}

@inproceedings{nickel2017poincare,
 author = {Nickel, Maximillian and Kiela, Douwe},
 booktitle = {Conference on Neural Information Processing Systems (NeurIPS)},
 pages = {6338--6347},
 title = {Poincar{\'e} embeddings for learning hierarchical representations},
 year = {2017}
}

@inproceedings{nickel2018learning,
 author = {Nickel, Maximillian and Kiela, Douwe},
 booktitle = {International Conference on Machine Learning (ICML)},
 pages = {3779--3788},
 title = {Learning continuous hierarchies in the Lorentz model of hyperbolic geometry},
 year = {2018}
}

@misc{openai2022,
 author = {OpenAI Foundation},
 howpublished = {\url{https://openai.com/index/chatgpt}},
 month = {November},
 title = {{Introducing chatgpt}},
 year = {2022}
}

@article{pal2024compositional,
 author = {Pal, Avik and van Spengler, Max and di Melendugno, Guido Maria D'Amely and Flaborea, Alessandro and Galasso, Fabio and Mettes, Pascal},
 journal = {arXiv preprint arXiv:2410.06912},
 title = {Compositional entailment learning for hyperbolic vision-language models},
 year = {2024}
}

@inproceedings{papadopoulos2010greedy,
 author = {Papadopoulos, Fragkiskos and Krioukov, Dmitri and Bogun{\'a}, Mari{\'a}n and Vahdat, Amin},
 booktitle = {Infocom},
 organization = {IEEE},
 pages = {1--9},
 title = {Greedy forwarding in dynamic scale-free networks embedded in hyperbolic metric spaces},
 year = {2010}
}

@article{patel2021nlp,
 author = {Patel, Arkil and Bhattamishra, Satwik and Goyal, Navin},
 journal = {arXiv preprint arXiv:2103.07191},
 title = {Are NLP models really able to solve simple math word problems?},
 year = {2021}
}

@article{peng2021hyperbolic,
 author = {Peng, Wei and Varanka, Tuomas and Mostafa, Abdelrahman and Shi, Henglin and Zhao, Guoying},
 journal = {IEEE Transactions on Pattern Analysis and Machine Intelligence},
 title = {Hyperbolic deep neural networks: A survey},
 year = {2021}
}

@inproceedings{poppi2025hyperbolic,
 author = {Tobia Poppi and Tejaswi Kasarla and Pascal Mettes and Lorenzo Baraldi and Rita Cucchiara},
 booktitle = {IEEE/CVF Conference on Computer Vision and Pattern Recognition (CVPR)},
 title = {Hyperbolic safety-aware vision-language models},
 url = {https://arxiv.org/abs/2503.12127},
 year = {2025}
}

@article{puccetti2022outliers,
 author = {Puccetti, Giovanni and Rogers, Anna and Drozd, Aleksandr and Dell'Orletta, Felice},
 journal = {arXiv preprint arXiv:2205.11380},
 title = {Outliers dimensions that disrupt transformers are driven by frequency},
 year = {2022}
}

@article{qin2021exploring,
 author = {Qin, Yujia and Wang, Xiaozhi and Su, Yusheng and Lin, Yankai and Ding, Ning and Yi, Jing and Chen, Weize and Liu, Zhiyuan and Li, Juanzi and Hou, Lei and others},
 journal = {arXiv preprint arXiv:2110.07867},
 title = {Exploring universal intrinsic task subspace via prompt tuning},
 year = {2021}
}

@article{qin2023chatgpt,
 author = {Qin, Chengwei and Zhang, Aston and Zhang, Zhuosheng and Chen, Jiaao and Yasunaga, Michihiro and Yang, Diyi},
 journal = {arXiv preprint arXiv:2302.06476},
 title = {Is chatgpt a general-purpose natural language processing task solver?},
 year = {2023}
}

@inproceedings{radford2021learning,
 author = {Alec Radford and Jong Wook Kim and Chris Hallacy and Aditya Ramesh and Gabriel Goh and Sandhini Agarwal and Girish Sastry and Amanda Askell and Pamela Mishkin and Jack Clark and Gretchen Krueger and Ilya Sutskever},
 booktitle = {International Conference on Machine Learning (ICML)},
 title = {Learning transferable visual models from natural language supervision},
 url = {https://arxiv.org/abs/2103.00020},
 year = {2021}
}

@inproceedings{rajbhandari2020zero,
 author = {Rajbhandari, Samyam and Rasley, Jeff and Ruwase, Olatunji and He, Yuxiong},
 booktitle = {SC20: International Conference for High Performance Computing, Networking, Storage and Analysis},
 organization = {IEEE},
 pages = {1--16},
 title = {Zero: Memory optimizations toward training trillion parameter models},
 year = {2020}
}

@article{ravasz2003hierarchical,
 author = {Ravasz, Erzs{\'e}bet and Barab{\'a}si, Albert-L{\'a}szl{\'o}},
 journal = {Physical review E},
 number = {2},
 pages = {026112},
 title = {Hierarchical organization in complex networks},
 volume = {67},
 year = {2003}
}

@inproceedings{reif2019visualizing,
 author = {Reif, Emily and Yuan, Ann and Wattenberg, Martin and Viegas, Fernanda B and Coenen, Andy and Pearce, Adam and Kim, Been},
 booktitle = {Conference on Neural Information Processing Systems (NeurIPS)},
 title = {Visualizing and measuring the geometry of BERT},
 volume = {32},
 year = {2019}
}

@article{ren2024mini,
 author = {Ren, Pengjie and Shi, Chengshun and Wu, Shiguang and Zhang, Mengqi and Ren, Zhaochun and de Rijke, Maarten and Chen, Zhumin and Pei, Jiahuan},
 journal = {arXiv preprint arXiv:2402.17263},
 title = {Mini-ensemble low-rank adapters for parameter-efficient fine-tuning},
 year = {2024}
}

@article{rudman2021isoscore,
 author = {Rudman, William and Gillman, Nate and Rayne, Taylor and Eickhoff, Carsten},
 journal = {arXiv preprint arXiv:2108.07344},
 title = {IsoScore: Measuring the uniformity of embedding space utilization},
 year = {2021}
}

@inproceedings{sarkar2011low,
 author = {Sarkar, Rik},
 booktitle = {International Symposium on Graph Drawing},
 organization = {Springer},
 pages = {355--366},
 title = {Low distortion delaunay embedding of trees in hyperbolic plane},
 year = {2011}
}

@article{sen2008collective,
 author = {Sen, Prithviraj and Namata, Galileo and Bilgic, Mustafa and Getoor, Lise and Galligher, Brian and Eliassi-Rad, Tina},
 journal = {AI magazine},
 number = {3},
 pages = {93--93},
 title = {Collective classification in network data},
 volume = {29},
 year = {2008}
}

@inproceedings{shen2024hugginggpt,
 author = {Shen, Yongliang and Song, Kaitao and Tan, Xu and Li, Dongsheng and Lu, Weiming and Zhuang, Yueting},
 booktitle = {Conference on Neural Information Processing Systems (NeurIPS)},
 title = {{HuggingGPT: Solving AI tasks with ChatGPT and its friends in Hugging Face}},
 volume = {36},
 year = {2024}
}

@article{shimizu2020hyperbolic,
 author = {Shimizu, Ryohei and Mukuta, Yusuke and Harada, Tatsuya},
 journal = {arXiv preprint arXiv:2006.08210},
 title = {Hyperbolic neural networks++},
 year = {2020}
}

@article{skopek2019mixed,
 author = {Skopek, Ondrej and Ganea, Octavian-Eugen and B{\'e}cigneul, Gary},
 journal = {arXiv preprint arXiv:1911.08411},
 title = {Mixed-curvature variational autoencoders},
 year = {2019}
}

@article{smith2014optimization,
 author = {Smith, Steven Thomas},
 journal = {arXiv preprint arXiv:1407.5965},
 title = {Optimization techniques on Riemannian manifolds},
 year = {2014}
}

@inproceedings{sun2021hgcf,
 author = {Sun, Jianing and Cheng, Zhaoyue and Zuberi, Saba and P{\'e}rez, Felipe and Volkovs, Maksims},
 booktitle = {The Web Conference (WWW)},
 pages = {593--601},
 title = {{HGCF}: Hyperbolic graph convolution networks for collaborative filtering},
 year = {2021}
}

@inproceedings{suzuki2021generalization,
 author = {Suzuki, Atsushi and Nitanda, Atsushi and Wang, Jing and Xu, Linchuan and Yamanishi, Kenji and Cavazza, Marc},
 booktitle = {International Conference on Machine Learning (ICML)},
 organization = {PMLR},
 pages = {10011--10021},
 title = {Generalization error bound for hyperbolic ordinal embedding},
 year = {2021}
}

@article{touvron2023llama,
 author = {Touvron, Hugo and Lavril, Thibaut and Izacard, Gautier and Martinet, Xavier and Lachaux, Marie-Anne and Lacroix, Timoth{\'e}e and Rozi{\`e}re, Baptiste and Goyal, Naman and Hambro, Eric and Azhar, Faisal and others},
 journal = {arXiv preprint arXiv:2302.13971},
 title = {Llama: Open and efficient foundation language models},
 year = {2023}
}

@inproceedings{van2023poincare,
 author = {van Spengler, Max and Berkhout, Erwin and Mettes, Pascal},
 booktitle = {IEEE/CVF International Conference on Computer Vision (ICCV)},
 pages = {5419--5428},
 title = {Poincar{\'e} resnet},
 year = {2023}
}

@article{wang2023lora,
 author = {Wang, Xi and Aitchison, Laurence and Rudolph, Maja},
 journal = {arXiv preprint arXiv:2310.00035},
 title = {Lora ensembles for large language model fine-tuning},
 year = {2023}
}

@inproceedings{weng2021unsupervised,
 author = {Weng, Zhenzhen and Ogut, Mehmet Giray and Limonchik, Shai and Yeung, Serena},
 booktitle = {IEEE/CVF Conference on Computer Vision and Pattern Recognition (CVPR)},
 pages = {2603--2612},
 title = {Unsupervised discovery of the long-tail in instance segmentation using hierarchical self-supervision},
 year = {2021}
}

@inproceedings{xiong2022hyperbolic,
 author = {Xiong, Bo and Cochez, Michael and Nayyeri, Mojtaba and Staab, Steffen},
 booktitle = {Conference on Neural Information Processing Systems (NeurIPS)},
 pages = {33016--33028},
 title = {Hyperbolic embedding inference for structured multi-label prediction},
 volume = {35},
 year = {2022}
}

@article{xu2023qa,
 author = {Xu, Yuhui and Xie, Lingxi and Gu, Xiaotao and Chen, Xin and Chang, Heng and Zhang, Hengheng and Chen, Zhensu and Zhang, Xiaopeng and Tian, Qi},
 journal = {arXiv preprint arXiv:2309.14717},
 title = {Qa-lora: Quantization-aware low-rank adaptation of large language models},
 year = {2023}
}

@inproceedings{yang2021discrete,
 author = {Yang, Menglin and Zhou, Min and Kalander, Marcus and Huang, Zengfeng and King, Irwin},
 booktitle = {ACM SIGKDD Conference on Knowledge Discovery and Data Mining (KDD)},
 pages = {1975--1985},
 title = {Discrete-time temporal network embedding via implicit hierarchical learning in hyperbolic space},
 year = {2021}
}

@inproceedings{yang2022hicf,
 author = {Yang, Menglin and Li, Zhihao and Zhou, Min and Liu, Jiahong and King, Irwin},
 booktitle = {ACM SIGKDD Conference on Knowledge Discovery and Data Mining (KDD)},
 pages = {2212--2221},
 title = {Hicf: Hyperbolic informative collaborative filtering},
 year = {2022}
}

@article{yang2022hyperbolic,
 author = {Yang, Menglin and Zhou, Min and Li, Zhihao and Liu, Jiahong and Pan, Lujia and Xiong, Hui and King, Irwin},
 journal = {arXiv preprint arXiv:2202.13852},
 title = {Hyperbolic graph neural networks: A review of methods and applications},
 year = {2022}
}

@article{yang2022hyperbolichtgn,
 author = {Yang, Menglin and Zhou, Min and Xiong, Hui and King, Irwin},
 journal = {IEEE Transactions on Knowledge and Data Engineering (TKDE)},
 title = {Hyperbolic temporal network embedding},
 year = {2022}
}

@inproceedings{yang2023kappahgcn,
 author = {Yang, Menglin and Zhou, Min and Pan, Lujia and King, Irwin},
 booktitle = {ACM SIGKDD Conference on Knowledge Discovery and Data Mining (KDD)},
 pages = {2965--2977},
 title = {$\kappa$hgcn: Tree-likeness modeling via continuous and discrete curvature learning},
 year = {2023}
}

@article{yang2024hypformer,
 author = {Yang, Menglin and Verma, Harshit and Zhang, Delvin Ce and Liu, Jiahong and King, Irwin and Ying, Rex},
 journal = {arXiv preprint arXiv:2407.01290},
 title = {Hypformer: Exploring efficient transformer fully in hyperbolic space},
 year = {2024}
}

@article{yang2024qwen2,
 author = {Yang, An and Yang, Baosong and Zhang, Beichen and Hui, Binyuan and Zheng, Bo and Yu, Bowen and Li, Chengyuan and Liu, Dayiheng and Huang, Fei and Wei, Haoran and others},
 journal = {arXiv preprint arXiv:2412.15115},
 title = {Qwen2. 5 technical report},
 year = {2024}
}

@inproceedings{zhang2023adaptive,
 author = {Zhang, Qingru and Chen, Minshuo and Bukharin, Alexander and He, Pengcheng and Cheng, Yu and Chen, Weizhu and Zhao, Tuo},
 booktitle = {International Conference on Learning Representations (ICLR)},
 title = {Adaptive budget allocation for parameter-efficient fine-tuning},
 year = {2023}
}

@article{zhu2021counter,
 author = {Zhu, Yaoming and Feng, Jiangtao and Zhao, Chengqi and Wang, Mingxuan and Li, Lei},
 journal = {arXiv preprint arXiv:2104.08154},
 title = {Counter-interference adapter for multilingual machine translation},
 year = {2021}
}

@book{ratcliffe2006foundations,
  title={Foundations of hyperbolic manifolds},
  author={Ratcliffe, John G},
  volume={149},
  year={2006},
  publisher={Springer}
}

@article{kipf2016semi,
  title={Semi-supervised classification with graph convolutional networks},
  author={Kipf, TN},
  journal={arXiv preprint arXiv:1609.02907},
  year={2016}
}

@inproceedings{sennrich2016neural,
  title={Neural machine translation of rare words with subword units},
  author={Sennrich, Rico and Haddow, Barry and Birch, Alexandra},
  booktitle={Proceedings of the 54th annual meeting of the association for computational linguistics (ACL)},
  pages={1715--1725},
  year={2016}
}
\newpage
\section*{NeurIPS Paper Checklist}

\begin{enumerate}

\item {\bf Claims}
    \item[] Question: Do the main claims made in the abstract and introduction accurately reflect the paper's contributions and scope?
    \item[] Answer: \answerYes{}
    \item[] Justification: The paper's abstract and introduction explicitly state the main claims, including the contribution of the proposed method and analysis, and its scope.
    \item[] Guidelines:
    \begin{itemize}
        \item The answer NA means that the abstract and introduction do not include the claims made in the paper.
        \item The abstract and/or introduction should clearly state the claims made, including the contributions made in the paper and important assumptions and limitations. A No or NA answer to this question will not be perceived well by the reviewers. 
        \item The claims made should match theoretical and experimental results, and reflect how much the results can be expected to generalize to other settings. 
        \item It is fine to include aspirational goals as motivation as long as it is clear that these goals are not attained by the paper. 
    \end{itemize}

\item {\bf Limitations}
    \item[] Question: Does the paper discuss the limitations of the work performed by the authors?
    \item[] Answer: \answerYes{}
    \item[] Justification:  The paper discusses a limitation in the Conclusion (Section 6).
    \item[] Guidelines:
    \begin{itemize}
        \item The answer NA means that the paper has no limitation while the answer No means that the paper has limitations, but those are not discussed in the paper. 
        \item The authors are encouraged to create a separate ``Limitations'' section in their paper.
        \item The paper should point out any strong assumptions and how robust the results are to violations of these assumptions (e.g., independence assumptions, noiseless settings, model well-specification, asymptotic approximations only holding locally). The authors should reflect on how these assumptions might be violated in practice and what the implications would be.
        \item The authors should reflect on the scope of the claims made, e.g., if the approach was only tested on a few datasets or with a few runs. In general, empirical results often depend on implicit assumptions, which should be articulated.
        \item The authors should reflect on the factors that influence the performance of the approach. For example, a facial recognition algorithm may perform poorly when image resolution is low or images are taken in low lighting. Or a speech-to-text system might not be used reliably to provide closed captions for online lectures because it fails to handle technical jargon.
        \item The authors should discuss the computational efficiency of the proposed algorithms and how they scale with dataset size.
        \item If applicable, the authors should discuss possible limitations of their approach to address problems of privacy and fairness.
        \item While the authors might fear that complete honesty about limitations might be used by reviewers as grounds for rejection, a worse outcome might be that reviewers discover limitations that aren't acknowledged in the paper. The authors should use their best judgment and recognize that individual actions in favor of transparency play an important role in developing norms that preserve the integrity of the community. Reviewers will be specifically instructed to not penalize honesty concerning limitations.
    \end{itemize}

\item {\bf Theory assumptions and proofs}
    \item[] Question: For each theoretical result, does the paper provide the full set of assumptions and a complete (and correct) proof?
    \item[] Answer: \answerYes{}
    \item[] Justification: The paper presents Proposition 1 regarding HypLoRA, introducing higher-order terms.  It states that the details and derivation of these terms are provided in Appendix F.  
    \item[] Guidelines:
    \begin{itemize}
        \item The answer NA means that the paper does not include theoretical results. 
        \item All theorems, formulas, and proofs in the paper should be numbered and cross-referenced.
        \item All assumptions should be clearly stated or referenced in the statement of any theorems.
        \item The proofs can either appear in the main paper or the supplemental material, but if they appear in the supplemental material, the authors are encouraged to provide a short proof sketch to provide intuition. 
        \item Inversely, any informal proof provided in the core of the paper should be complemented by formal proofs provided in appendix or supplemental material.
        \item Theorems and Lemmas that the proof relies upon should be properly referenced. 
    \end{itemize}

    \item {\bf Experimental result reproducibility}
    \item[] Question: Does the paper fully disclose all the information needed to reproduce the main experimental results of the paper to the extent that it affects the main claims and/or conclusions of the paper (regardless of whether the code and data are provided or not)?
    \item[] Answer: \answerYes{}
    \item[] Justification: Section 5.1 ``Experimental Settings'' describes the datasets, models, and implementation details, including optimizer (AdamW), etc.
    \item[] Guidelines:
    \begin{itemize}
        \item The answer NA means that the paper does not include experiments.
        \item If the paper includes experiments, a No answer to this question will not be perceived well by the reviewers: Making the paper reproducible is important, regardless of whether the code and data are provided or not.
        \item If the contribution is a dataset and/or model, the authors should describe the steps taken to make their results reproducible or verifiable. 
        \item Depending on the contribution, reproducibility can be accomplished in various ways. For example, if the contribution is a novel architecture, describing the architecture fully might suffice, or if the contribution is a specific model and empirical evaluation, it may be necessary to either make it possible for others to replicate the model with the same dataset, or provide access to the model. In general. releasing code and data is often one good way to accomplish this, but reproducibility can also be provided via detailed instructions for how to replicate the results, access to a hosted model (e.g., in the case of a large language model), releasing of a model checkpoint, or other means that are appropriate to the research performed.
        \item While NeurIPS does not require releasing code, the conference does require all submissions to provide some reasonable avenue for reproducibility, which may depend on the nature of the contribution. For example
        \begin{enumerate}
            \item If the contribution is primarily a new algorithm, the paper should make it clear how to reproduce that algorithm.
            \item If the contribution is primarily a new model architecture, the paper should describe the architecture clearly and fully.
            \item If the contribution is a new model (e.g., a large language model), then there should either be a way to access this model for reproducing the results or a way to reproduce the model (e.g., with an open-source dataset or instructions for how to construct the dataset).
            \item We recognize that reproducibility may be tricky in some cases, in which case authors are welcome to describe the particular way they provide for reproducibility. In the case of closed-source models, it may be that access to the model is limited in some way (e.g., to registered users), but it should be possible for other researchers to have some path to reproducing or verifying the results.
        \end{enumerate}
    \end{itemize}

\item {\bf Open access to data and code}
    \item[] Question: Does the paper provide open access to the data and code, with sufficient instructions to faithfully reproduce the main experimental results, as described in supplemental material?
    \item[] Answer: \answerYes{}
    \item[] Justification: The datasets used (e.g., GSM8K, AQUA, various commonsense reasoning benchmarks) are publicly available and cited.  
    \item[] Guidelines:
    \begin{itemize}
        \item The answer NA means that paper does not include experiments requiring code.
        \item Please see the NeurIPS code and data submission guidelines (\url{https://nips.cc/public/guides/CodeSubmissionPolicy}) for more details.
        \item While we encourage the release of code and data, we understand that this might not be possible, so “No” is an acceptable answer. Papers cannot be rejected simply for not including code, unless this is central to the contribution (e.g., for a new open-source benchmark).
        \item The instructions should contain the exact command and environment needed to run to reproduce the results. See the NeurIPS code and data submission guidelines (\url{https://nips.cc/public/guides/CodeSubmissionPolicy}) for more details.
        \item The authors should provide instructions on data access and preparation, including how to access the raw data, preprocessed data, intermediate data, and generated data, etc.
        \item The authors should provide scripts to reproduce all experimental results for the new proposed method and baselines. If only a subset of experiments are reproducible, they should state which ones are omitted from the script and why.
        \item At submission time, to preserve anonymity, the authors should release anonymized versions (if applicable).
        \item Providing as much information as possible in supplemental material (appended to the paper) is recommended, but including URLs to data and code is permitted.
    \end{itemize}

\item {\bf Experimental setting/details}
    \item[] Question: Does the paper specify all the training and test details (e.g., data splits, hyperparameters, how they were chosen, type of optimizer, etc.) necessary to understand the results?
    \item[] Answer: \answerYes{}
    \item[] Justification: Section 5.1 details the experimental setup.
    \item[] Guidelines:
    \begin{itemize}
        \item The answer NA means that the paper does not include experiments.
        \item The experimental setting should be presented in the core of the paper to a level of detail that is necessary to appreciate the results and make sense of them.
        \item The full details can be provided either with the code, in appendix, or as supplemental material.
    \end{itemize}

\item {\bf Experiment statistical significance}
    \item[] Question: Does the paper report error bars suitably and correctly defined or other appropriate information about the statistical significance of the experiments?
    \item[] Answer: \answerYes{}
    \item[] Justification: Table~\ref{tab:hyperbolicity_table}, which presents hyperbolicity across various metric spaces and datasets, includes error bars reported as mean and standard deviation. Tables 3 and 4 report mean accuracies over three runs.
    \item[] Guidelines:
    \begin{itemize}
        \item The answer NA means that the paper does not include experiments.
        \item The authors should answer ``Yes'' if the results are accompanied by error bars, confidence intervals, or statistical significance tests, at least for the experiments that support the main claims of the paper.
        \item The factors of variability that the error bars are capturing should be clearly stated (for example, train/test split, initialization, random drawing of some parameter, or overall run with given experimental conditions).
        \item The method for calculating the error bars should be explained (closed form formula, call to a library function, bootstrap, etc.)
        \item The assumptions made should be given (e.g., Normally distributed errors).
        \item It should be clear whether the error bar is the standard deviation or the standard error of the mean.
        \item It is OK to report 1-sigma error bars, but one should state it. The authors should preferably report a 2-sigma error bar than state that they have a 96\% CI, if the hypothesis of Normality of errors is not verified.
        \item For asymmetric distributions, the authors should be careful not to show in tables or figures symmetric error bars that would yield results that are out of range (e.g. negative error rates).
        \item If error bars are reported in tables or plots, The authors should explain in the text how they were calculated and reference the corresponding figures or tables in the text.
    \end{itemize}

\item {\bf Experiments compute resources}
    \item[] Question: For each experiment, does the paper provide sufficient information on the computer resources (type of compute workers, memory, time of execution) needed to reproduce the experiments?
    \item[] Answer: \answerYes{}
    \item[] Justification: The paper provides implementation details in Section 5.1, including the GPU type (NVIDIA A100) used for experiments.
    \item[] Guidelines:
    \begin{itemize}
        \item The answer NA means that the paper does not include experiments.
        \item The paper should indicate the type of compute workers CPU or GPU, internal cluster, or cloud provider, including relevant memory and storage.
        \item The paper should provide the amount of compute required for each of the individual experimental runs as well as estimate the total compute. 
        \item The paper should disclose whether the full research project required more compute than the experiments reported in the paper (e.g., preliminary or failed experiments that didn't make it into the paper). 
    \end{itemize}
    
\item {\bf Code of ethics}
    \item[] Question: Does the research conducted in the paper conform, in every respect, with the NeurIPS Code of Ethics \url{https://neurips.cc/public/EthicsGuidelines}?
    \item[] Answer: \answerYes{}
    \item[] Justification: Based on the content of the paper, the research focuses on investigating geometric properties of LLMs and proposing a new fine-tuning method. The datasets used are standard benchmarks in the field.  There is no indication of activities that would violate the NeurIPS Code of Ethics, such as plagiarism, falsification of data, or unethical use of human subjects. 
    \item[] Guidelines:
    \begin{itemize}
        \item The answer NA means that the authors have not reviewed the NeurIPS Code of Ethics.
        \item If the authors answer No, they should explain the special circumstances that require a deviation from the Code of Ethics.
        \item The authors should make sure to preserve anonymity (e.g., if there is a special consideration due to laws or regulations in their jurisdiction).
    \end{itemize}

\item {\bf Broader impacts}
    \item[] Question: Does the paper discuss both potential positive societal impacts and negative societal impacts of the work performed?
    \item[] Answer: \answerYes{}
    \item[] Justification: Conclusion part.
    \item[] Guidelines:
    \begin{itemize}
        \item The answer NA means that there is no societal impact of the work performed.
        \item If the authors answer NA or No, they should explain why their work has no societal impact or why the paper does not address societal impact.
        \item Examples of negative societal impacts include potential malicious or unintended uses (e.g., disinformation, generating fake profiles, surveillance), fairness considerations (e.g., deployment of technologies that could make decisions that unfairly impact specific groups), privacy considerations, and security considerations.
        \item The conference expects that many papers will be foundational research and not tied to particular applications, let alone deployments. However, if there is a direct path to any negative applications, the authors should point it out. For example, it is legitimate to point out that an improvement in the quality of generative models could be used to generate deepfakes for disinformation. On the other hand, it is not needed to point out that a generic algorithm for optimizing neural networks could enable people to train models that generate Deepfakes faster.
        \item The authors should consider possible harms that could arise when the technology is being used as intended and functioning correctly, harms that could arise when the technology is being used as intended but gives incorrect results, and harms following from (intentional or unintentional) misuse of the technology.
        \item If there are negative societal impacts, the authors could also discuss possible mitigation strategies (e.g., gated release of models, providing defenses in addition to attacks, mechanisms for monitoring misuse, mechanisms to monitor how a system learns from feedback over time, improving the efficiency and accessibility of ML).
    \end{itemize}
    
\item {\bf Safeguards}
    \item[] Question: Does the paper describe safeguards that have been put in place for responsible release of data or models that have a high risk for misuse (e.g., pretrained language models, image generators, or scraped datasets)?
    \item[] Answer: \answerNA{}
    \item[] Justification: It does not introduce new large-scale pretrained models or novel scraped datasets that would pose a high risk for misuse requiring specific safeguards beyond those applicable to the original models it builds upon.
    \item[] Guidelines:
    \begin{itemize}
        \item The answer NA means that the paper poses no such risks.
        \item Released models that have a high risk for misuse or dual-use should be released with necessary safeguards to allow for controlled use of the model, for example by requiring that users adhere to usage guidelines or restrictions to access the model or implementing safety filters. 
        \item Datasets that have been scraped from the Internet could pose safety risks. The authors should describe how they avoided releasing unsafe images.
        \item We recognize that providing effective safeguards is challenging, and many papers do not require this, but we encourage authors to take this into account and make a best faith effort.
    \end{itemize}

\item {\bf Licenses for existing assets}
    \item[] Question: Are the creators or original owners of assets (e.g., code, data, models), used in the paper, properly credited and are the license and terms of use explicitly mentioned and properly respected?
    \item[] Answer: \answerYes{}
    \item[] Justification: The creators/original owners of assets (datasets like GSM8K, MAWPS, AQUA; models like LLaMA, Gemma; and software like the Powerlaw Package ) are credited via citations. 
    \item[] Guidelines:
    \begin{itemize}
        \item The answer NA means that the paper does not use existing assets.
        \item The authors should cite the original paper that produced the code package or dataset.
        \item The authors should state which version of the asset is used and, if possible, include a URL.
        \item The name of the license (e.g., CC-BY 4.0) should be included for each asset.
        \item For scraped data from a particular source (e.g., website), the copyright and terms of service of that source should be provided.
        \item If assets are released, the license, copyright information, and terms of use in the package should be provided. For popular datasets, \url{paperswithcode.com/datasets} has curated licenses for some datasets. Their licensing guide can help determine the license of a dataset.
        \item For existing datasets that are re-packaged, both the original license and the license of the derived asset (if it has changed) should be provided.
        \item If this information is not available online, the authors are encouraged to reach out to the asset's creators.
    \end{itemize}

\item {\bf New assets}
    \item[] Question: Are new assets introduced in the paper well documented and is the documentation provided alongside the assets?
    \item[] Answer: \answerNA{}
    \item[] Justification: No new assets.
    \item[] Guidelines:
    \begin{itemize}
        \item The answer NA means that the paper does not release new assets.
        \item Researchers should communicate the details of the dataset/code/model as part of their submissions via structured templates. This includes details about training, license, limitations, etc. 
        \item The paper should discuss whether and how consent was obtained from people whose asset is used.
        \item At submission time, remember to anonymize your assets (if applicable). You can either create an anonymized URL or include an anonymized zip file.
    \end{itemize}

\item {\bf Crowdsourcing and research with human subjects}
    \item[] Question: For crowdsourcing experiments and research with human subjects, does the paper include the full text of instructions given to participants and screenshots, if applicable, as well as details about compensation (if any)? 
    \item[] Answer: \answerNA{}
    \item[] Justification: The datasets are existing benchmarks.
    \item[] Guidelines:
    \begin{itemize}
        \item The answer NA means that the paper does not involve crowdsourcing nor research with human subjects.
        \item Including this information in the supplemental material is fine, but if the main contribution of the paper involves human subjects, then as much detail as possible should be included in the main paper. 
        \item According to the NeurIPS Code of Ethics, workers involved in data collection, curation, or other labor should be paid at least the minimum wage in the country of the data collector. 
    \end{itemize}

\item {\bf Institutional review board (IRB) approvals or equivalent for research with human subjects}
    \item[] Question: Does the paper describe potential risks incurred by study participants, whether such risks were disclosed to the subjects, and whether Institutional Review Board (IRB) approvals (or an equivalent approval/review based on the requirements of your country or institution) were obtained?
    \item[] Answer: \answerNA{}
    \item[] Justification: The research does not appear to involve human subjects in a way that would necessitate IRB approval, as per the justification for question 14.
    \item[] Guidelines:
    \begin{itemize}
        \item The answer NA means that the paper does not involve crowdsourcing nor research with human subjects.
        \item Depending on the country in which research is conducted, IRB approval (or equivalent) may be required for any human subjects research. If you obtained IRB approval, you should clearly state this in the paper. 
        \item We recognize that the procedures for this may vary significantly between institutions and locations, and we expect authors to adhere to the NeurIPS Code of Ethics and the guidelines for their institution. 
        \item For initial submissions, do not include any information that would break anonymity (if applicable), such as the institution conducting the review.
    \end{itemize}

\item {\bf Declaration of LLM usage}
    \item[] Question: Does the paper describe the usage of LLMs if it is an important, original, or non-standard component of the core methods in this research? Note that if the LLM is used only for writing, editing, or formatting purposes and does not impact the core methodology, scientific rigorousness, or originality of the research, declaration is not required.
    \item[] Answer: \answerNo{}
    \item[] Justification: No, except for writing, editing, or formatting purposes.
    \item[] Guidelines:
    \begin{itemize}
        \item The answer NA means that the core method development in this research does not involve LLMs as any important, original, or non-standard components.
        \item Please refer to our LLM policy (\url{https://neurips.cc/Conferences/2025/LLM}) for what should or should not be described.
    \end{itemize}

\end{enumerate}

\newpage
\appendix
\startcontents[sections]
\printcontents[sections]{}{0}
{\section*{Appendix Contents}\setcounter{tocdepth}{2}}
\newpage

\section{More Investigation Results}
\label{sec:appendix_more_investigation}

\subsection{Token Frequency and Norm Distribution on Mathematical Reasoning}

To provide a comprehensive understanding of the geometric properties of token embeddings across different mathematical reasoning tasks, we extend our analysis beyond the GSM8K dataset presented in the main text to include AQuA and MAWPS datasets. This broader investigation allows us to validate the consistency of our findings across diverse mathematical problem types and complexity levels. The AQuA dataset presents algebraic word problems that require multi-step reasoning and equation solving, while MAWPS focuses on elementary arithmetic word problems with varying structural complexity. By analyzing token distributions across these complementary datasets, we can assess whether the observed power-law behavior and hierarchical token organization represent universal properties of mathematical reasoning tasks or are specific to particular problem domains.

\begin{figure}[h]
   \centering
    \includegraphics[width=0.32\textwidth]{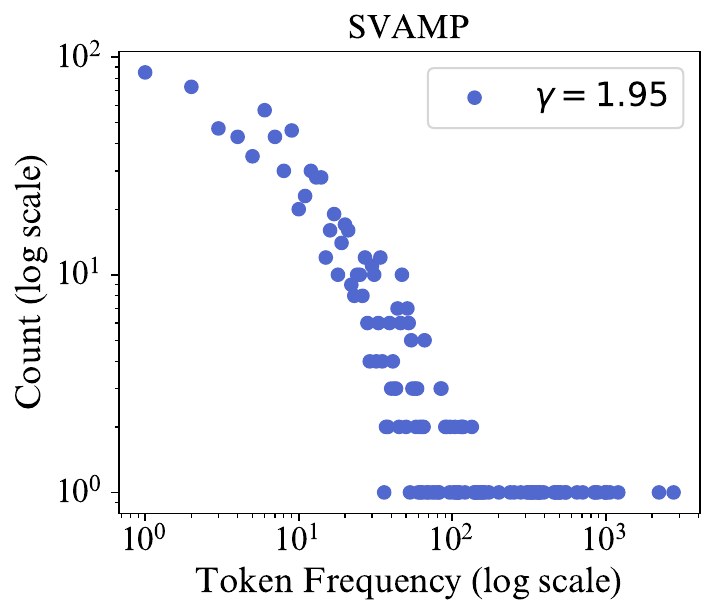}
        \includegraphics[width=0.32\textwidth]{figures/token_freq/AQuA.pdf}
            \includegraphics[width=0.32\textwidth]{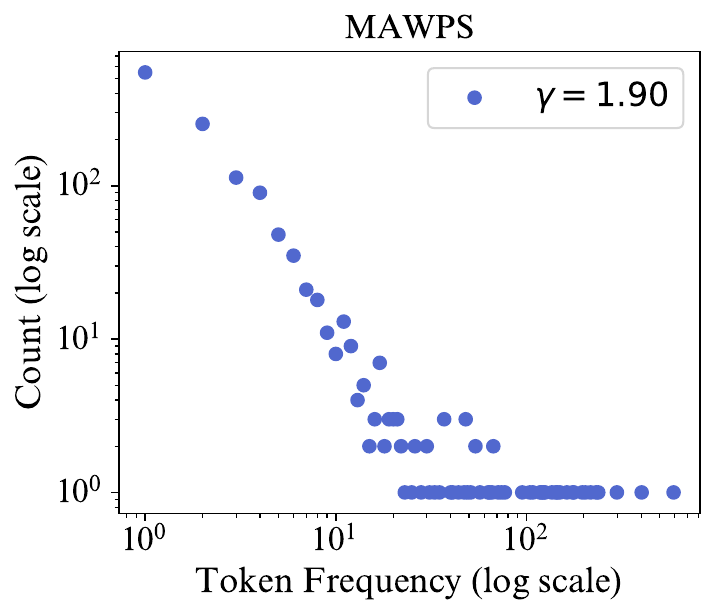}

    \includegraphics[width=0.32\textwidth]{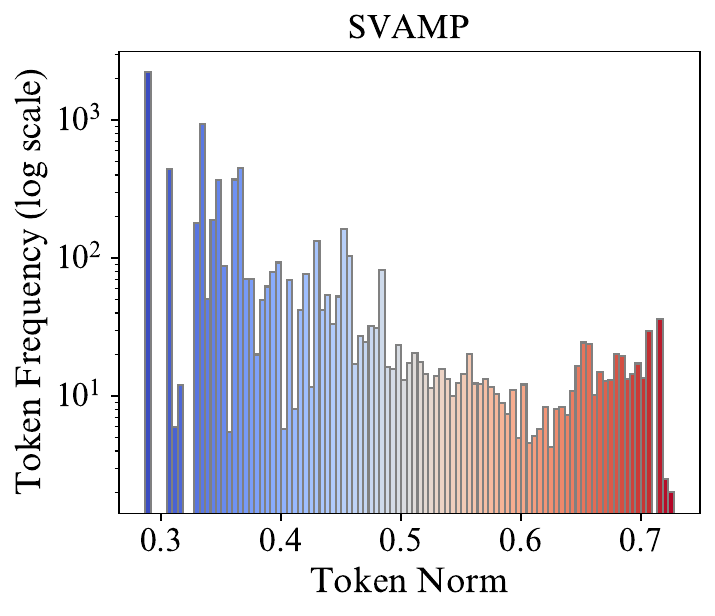}
        \includegraphics[width=0.32\textwidth]{figures/token_norm/AQuA.pdf}
            \includegraphics[width=0.32\textwidth]{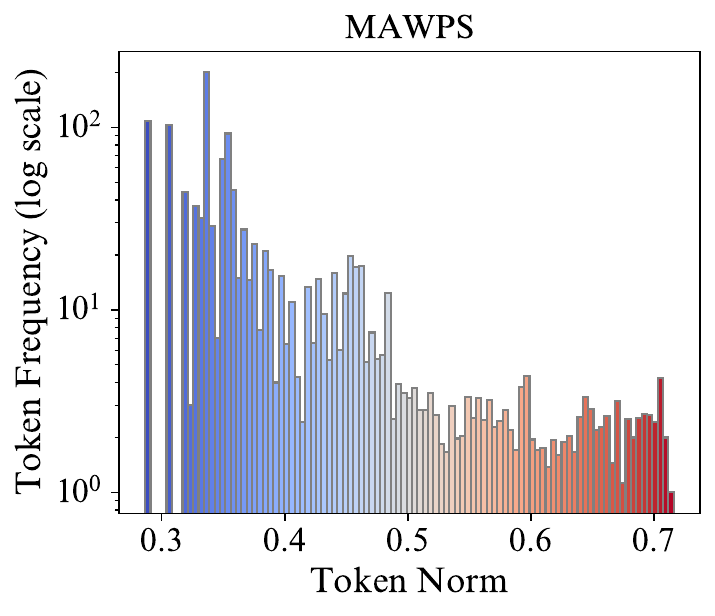}
   \caption{Token frequency distribution (top row) and token frequency vs.\ norm (bottom row) across different mathematical reasoning datasets in LLaMA3. The top row shows the power-law distribution of token frequencies with the decay rate ($\gamma$) annotated for each dataset. The bottom row illustrates the relationship between token frequency and token norm, binned and colored by frequency, where higher token norms correspond to lower frequencies.}
   \label{fig:appendix_token_frequency_distribution}
\end{figure}

\begin{figure}
   \centering
        \includegraphics[width=0.32\textwidth]{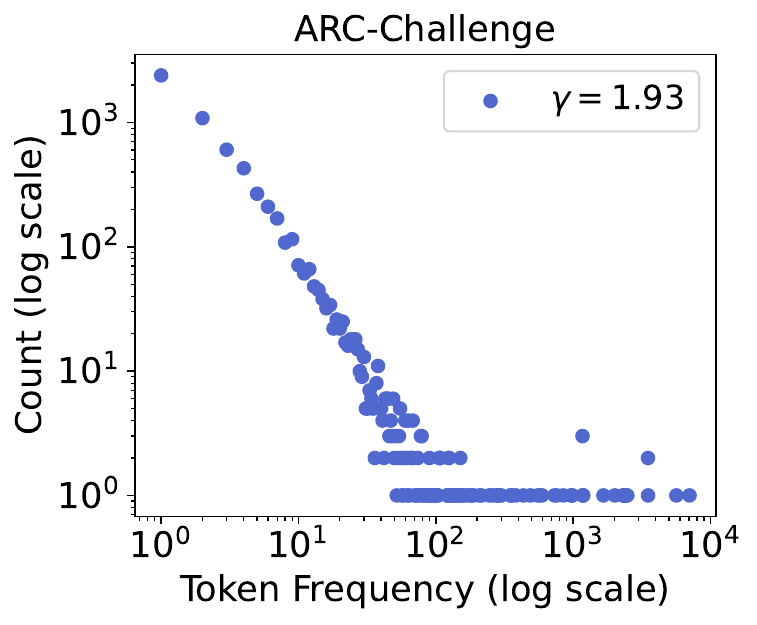}
        \includegraphics[width=0.32\textwidth]{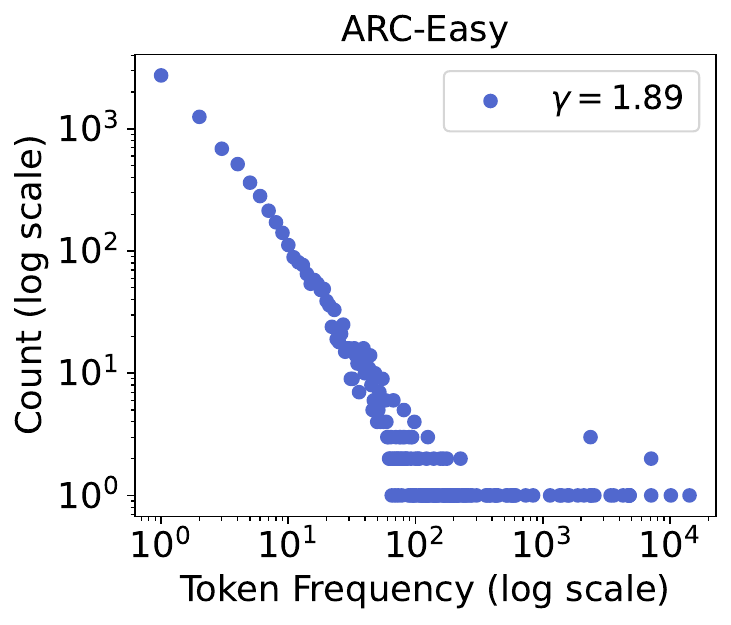}
            \includegraphics[width=0.32\textwidth]{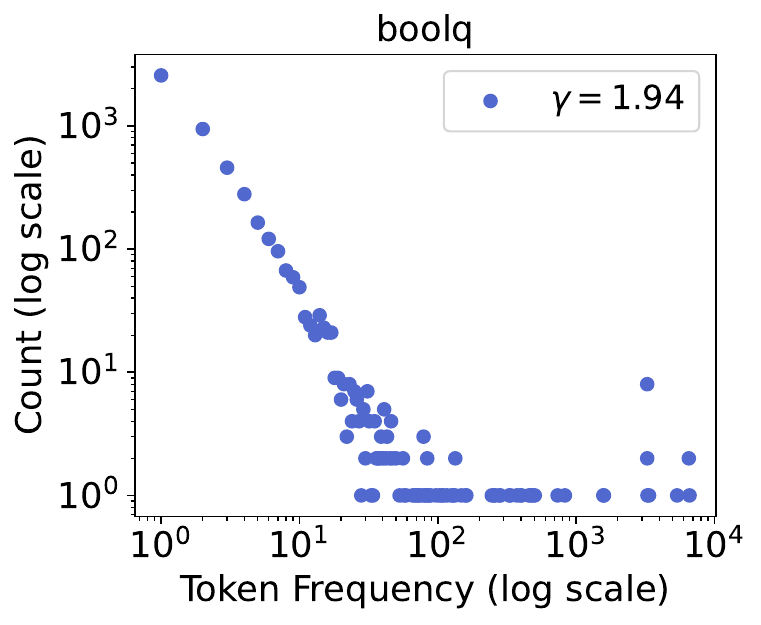}

                    \includegraphics[width=0.32\textwidth]{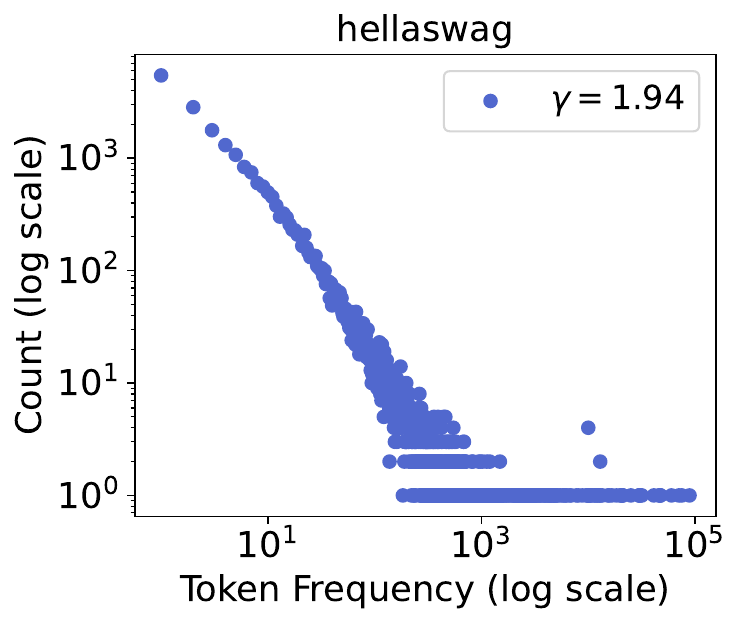}
        \includegraphics[width=0.32\textwidth]{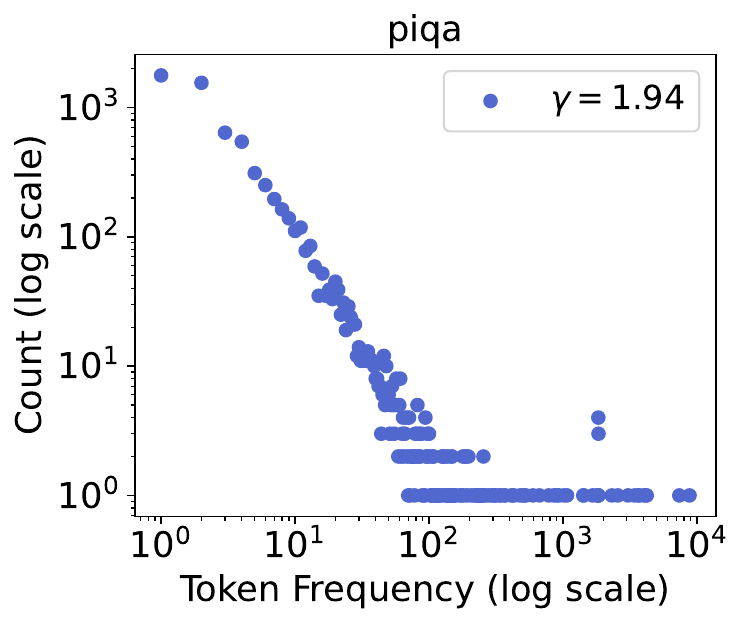}
            \includegraphics[width=0.32\textwidth]{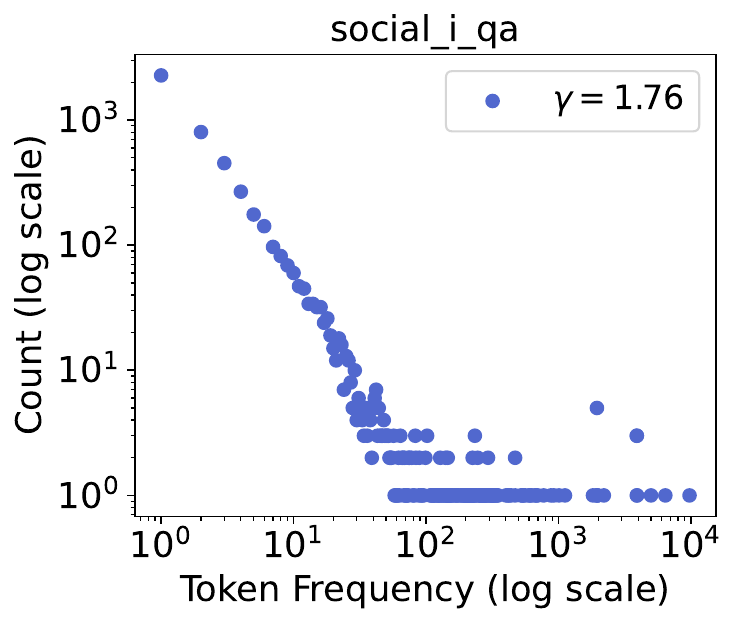}

            \includegraphics[width=0.32\textwidth]{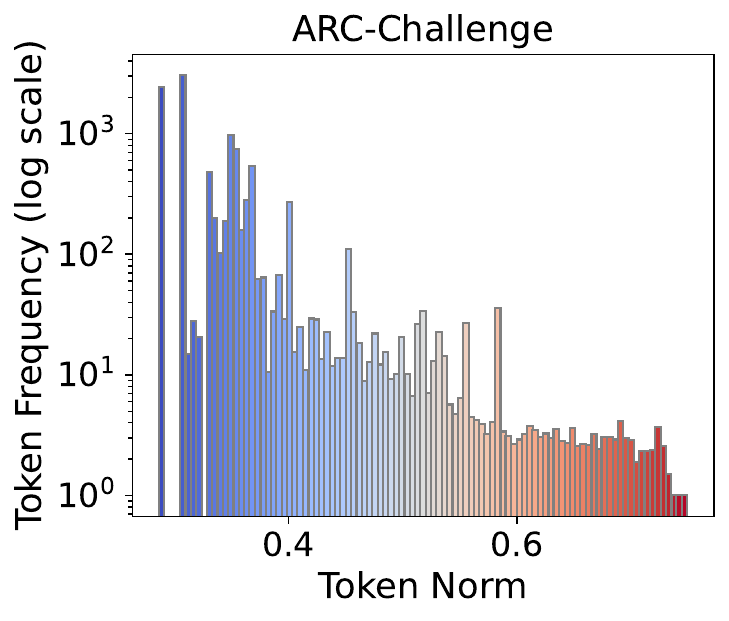}
        \includegraphics[width=0.32\textwidth]{figures/commonsense/LLaMA3_8B/token_norm_ARC_Challenge.pdf}
            \includegraphics[width=0.32\textwidth]{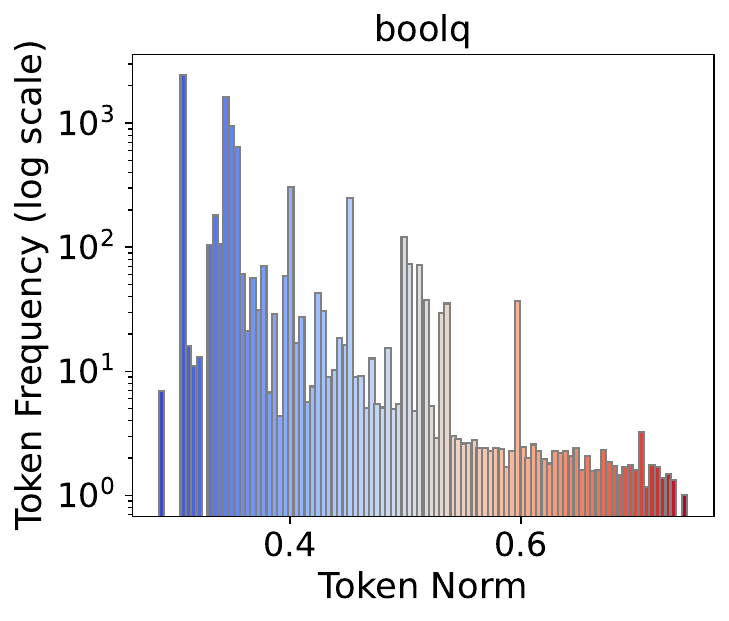}

                    \includegraphics[width=0.32\textwidth]{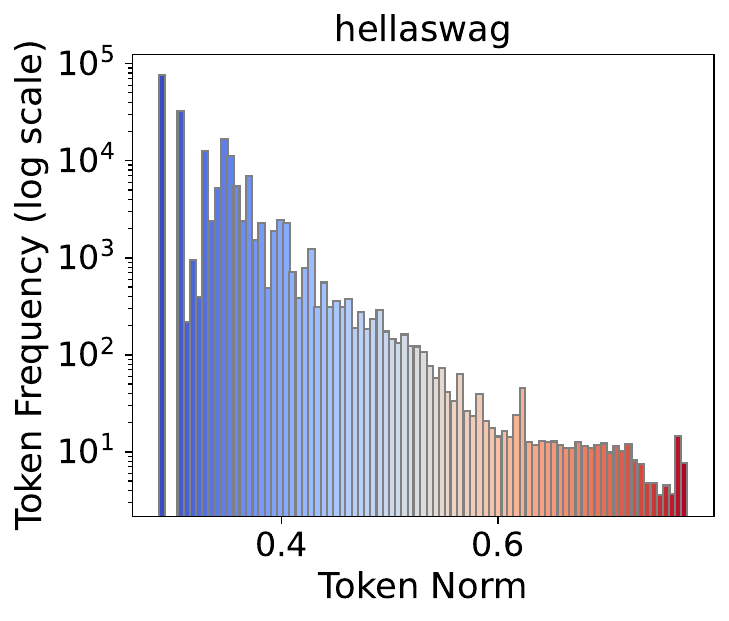}
        \includegraphics[width=0.32\textwidth]{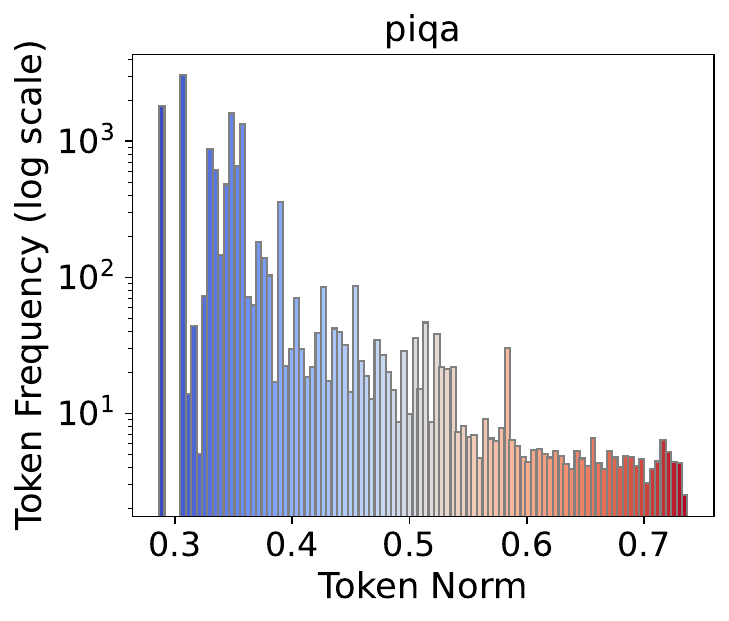}
            \includegraphics[width=0.32\textwidth]{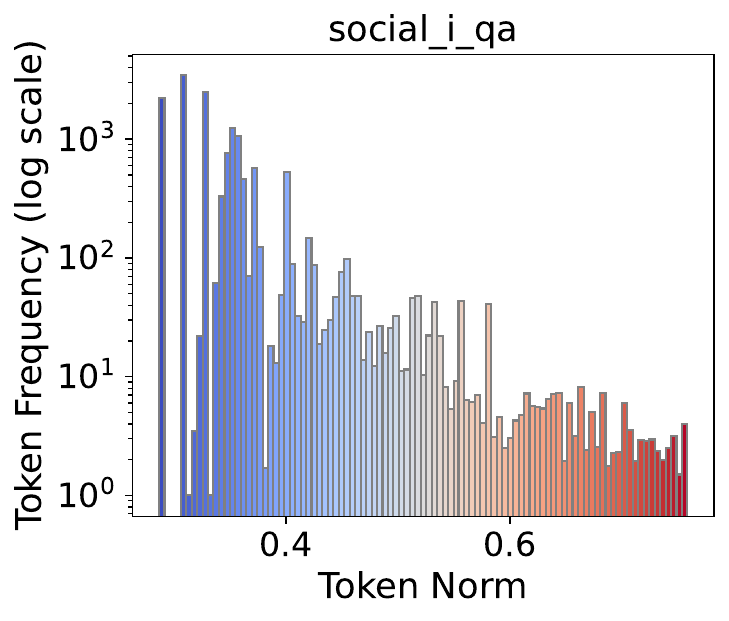}
   \caption{Token frequency distribution (top two rows) and token frequency vs. norm (bottom two rows) across different commonsense reasoning datasets in LLaMA3. The top two rows show the power-law distribution of token frequencies with the decay rate ($\gamma$) annotated for each dataset. The bottom two rows illustrate the relationship between token frequency and token norm, binned and colored by frequency, where higher token norms correspond to lower frequencies.}
   \label{fig:appendix_token_frequency_distribution_common_sense}
\end{figure}

Our extended analysis, illustrated in Figure~\ref{fig:appendix_token_frequency_distribution}, reveals remarkably consistent patterns across all three mathematical reasoning datasets. The power-law exponents remain stable within a narrow range ($\gamma \in [1.89, 1.95]$), indicating that the hierarchical structure of mathematical language is preserved regardless of the specific problem type or complexity level. The relationship between token frequency and embedding norms shows consistent inverse correlation across all datasets, with high-frequency mathematical operators and common function words clustering near the origin, while domain-specific mathematical terms and numerical values are positioned at greater distances. \textbf{This consistency strengthens our hypothesis that mathematical reasoning tasks inherently exhibit hyperbolic characteristics in their token embedding spaces}, providing strong empirical support for the effectiveness of hyperbolic fine-tuning approaches like \method in mathematical domains.

\subsection{Token Frequency and Norm Distribution on Commonsense Reasoning}
To demonstrate the generalizability of our findings beyond mathematical reasoning, we conduct a comprehensive analysis of token distributions across six diverse commonsense reasoning datasets: ARC-Challenge, ARC-Easy, BoolQ, HellaSwag, PIQA, and SIQA.
These datasets span a wide range of commonsense reasoning tasks, from factual knowledge retrieval (ARC datasets) and yes/no question answering (BoolQ) to physical commonsense (PIQA) and social understanding (SIQA). This diverse collection allows us to investigate whether the hyperbolic characteristics observed in mathematical reasoning extend to broader domains of human knowledge and reasoning. The inclusion of both challenging (ARC-Challenge, HellaSwag) and more accessible (ARC-Easy, BoolQ) datasets enables us to examine how task difficulty influences the underlying geometric structure of token embeddings.

The results presented in Figure~\ref{fig:appendix_token_frequency_distribution_common_sense} demonstrate that the power-law distribution of token frequencies and the inverse relationship between frequency and embedding norms persist across all commonsense reasoning datasets, with power-law exponents ranging from $\gamma = 1.76$ to $\gamma = 1.94$. Notably, the Social IQA dataset exhibits a slightly lower exponent $(\gamma = 1.76)$, suggesting that social reasoning tasks may have a somewhat different hierarchical structure, possibly due to the more nuanced and context-dependent nature of social interactions compared to factual or physical reasoning. Despite this variation, the overall pattern remains consistent: abstract concepts and function words maintain smaller norms and higher frequencies, while specific entities, proper nouns, and domain-specific terminology are positioned at greater distances from the origin. 

\begin{table}[t]
\caption{Relative $\delta$-hyperbolicity (mean $\pm$ std.) of the final hidden layer in Gemma-7B across math (AQuA, GSM8K) and commonsense (ARC-Challenge, WinoGrande, OpenBookQA) datasets comparing the frozen base model, LoRA, DoRA, and \method.}
\centering
\resizebox{0.7\textwidth}{!}{
\begin{tabular}{lcccc}
\toprule
\textbf{Dataset} & \textbf{Base Model} & \textbf{LoRA} & \textbf{DoRA} & \textbf{\method} \\
\midrule
AQuA         & $0.31 \pm 0.04$ & $0.24 \pm 0.05$ & $0.23 \pm 0.05$ & \cellcolor{lightgray}$0.22 \pm 0.03$ \\
GSM8K        & $0.28 \pm 0.04$ & $0.21 \pm 0.05$ & $0.21 \pm 0.05$ & \cellcolor{lightgray}$0.20 \pm 0.03$ \\
ARC-Challenge & $0.30 \pm 0.03$ & $0.35 \pm 0.03$ & $0.36 \pm 0.02$ & \cellcolor{lightgray}$0.25 \pm 0.02$ \\
Winogrande   & $0.22 \pm 0.04$ & $0.32 \pm 0.02$ & $0.27 \pm 0.02$ & \cellcolor{lightgray}$0.27 \pm 0.02$ \\
OpenbookQA   & $0.30 \pm 0.03$ & $0.35 \pm 0.03$ & $0.38 \pm 0.02$ & \cellcolor{lightgray}$0.25 \pm 0.02$ \\
\bottomrule
\end{tabular}
}
\label{tab:hyperbolicity_table_gemma7b_last_hidden}
\end{table}

\begin{table}[t]
\caption{Relative $\delta$-hyperbolicity (mean $\pm$ std.) of the final hidden layer in Gemma3-4B for the same five datasets, contrasting the base model with LoRA, DoRA, and \method.}
\centering
\resizebox{0.7\textwidth}{!}{
\begin{tabular}{lcccc}
\toprule
\textbf{Dataset} & \textbf{Base Model} & \textbf{LoRA} & \textbf{DoRA} & \textbf{\method} \\
\midrule
AQuA         & $0.17 \pm 0.03$ & $0.17 \pm 0.03$ & $0.19 \pm 0.02$ & \cellcolor{lightgray}$0.11 \pm 0.01$ \\
GSM8K        & $0.16 \pm 0.03$ & $0.20 \pm 0.03$ & $0.19 \pm 0.03$ & \cellcolor{lightgray}$0.11 \pm 0.02$ \\
ARC-Challenge & $0.17 \pm 0.02$ & $0.21 \pm 0.01$ & $0.17 \pm 0.02$ & \cellcolor{lightgray}$0.20 \pm 0.02$ \\
Winogrande   & $0.16 \pm 0.02$ & $0.16 \pm 0.02$ & $0.21 \pm 0.01$ & \cellcolor{lightgray}$0.12 \pm 0.01$ \\
OpenbookQA   & $0.17 \pm 0.03$ & $0.16 \pm 0.02$ & $0.17 \pm 0.03$ & \cellcolor{lightgray}$0.11 \pm 0.01$ \\
\bottomrule
\end{tabular}
}
\label{tab:hyperbolicity_table_gemma3b_last_hidden}
\end{table}

\subsection{Hyperbolicity in the Final Hidden Layer of LLMs}
\label{sec:hyperbolicity-final-layer}

In this part, we further present the analysis of the hyperbolicity of the hidden states in Tables~\ref{tab:hyperbolicity_table_gemma7b_last_hidden} and Table~\ref{tab:hyperbolicity_table_gemma3b_last_hidden}. Considering five distinct reasoning datasets, including two mathematical reasoning datasets (AQuA and GSM8K) as well as three commonsense reasoning datasets (ARC-Challenge, Winogrande, and OpenbookQA), we observe that the base models consistently exhibit less hyperbolic structure (i.e., higher $\delta$ values) in their final hidden layer representations compared to their initial token embeddings.

LoRA and DoRA generally reduce the $\delta$ values, while the proposed \method method mostly achieves even lower values, indicating a higher degree of hyperbolicity in the learned representations. This effect is observed across most datasets in both model families. These empirical findings complement our analysis of the initial token embeddings: while the pretrained models begin with a latent hierarchical structure, as evidenced by hyperbolicity in the input layer, fine-tuning methods can either preserve or distort this property. The consistently lower $\delta$ values of \method provide strong empirical evidence that our method actively preserves and enhances the hierarchical structure of the representations throughout the model, aligning the final contextualized embeddings with the geometric biases that are beneficial for reasoning.

\section{Hyperbolicity on Different Metric Spaces}
\label{sec:hyperbolicity_different_metric_space}
Table~\ref{tab:hyperbolicity_table} presents the hyperbolicity values in both continuous (i.e., sphere space) and discrete metric spaces (i.e., tree, scale-free, and random graphs). 
We employ a consistent processing method similar to that used in Section~\ref{sec:hyperbolicity_investigation} for embedding spaces. Specifically, we sample 1{,}000 four-tuples, compute the $\delta$ value for each, and then take the maximum value.

\begin{figure}[h]
    \centering
    \includegraphics[width=0.4\textwidth]{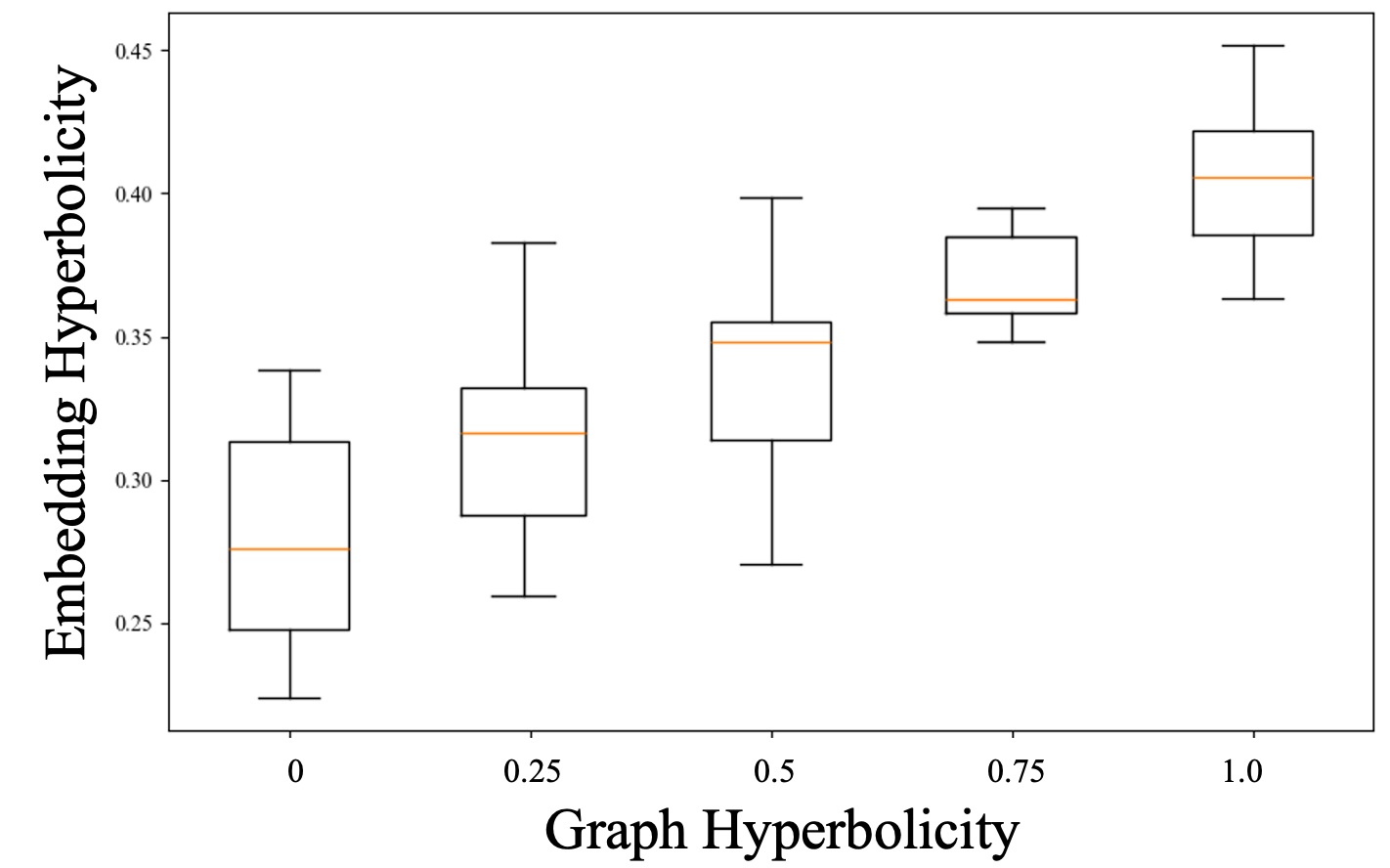}
    \caption{Empirical correlation between the ground-truth $\delta$-hyperbolicity of several reference graphs (tree, scale-free, PubMed, dense, sphere) and the $\delta$ measured after embedding them with a two-layer GCN into Euclidean space; each point averages 1{,}000 sampled quadruples.}
    \label{fig:embedding_hyperbolicity}
\end{figure}
For the sphere space, we use a two-dimensional model and calculate hyperbolicity based on geodesic distances. The PubMed graph is sourced from Sen et al.~\citep{sen2008collective}. 
The tree and dense graphs are generated using NetworkX~\citep{hagberg2008exploring}. For these graphs, we remove isolated nodes before performing our calculations to ensure consistency. 
We use the shortest-path distance on each graph as the distance measure, analogous to the concept of geodesics in continuous spaces.

In this study, we utilize the Euclidean distance to compute the hyperbolicity of token embeddings, following the approach proposed by~\cite{khrulkov2020hyperbolic}. To further assess the validity of this method, we embed graphs with varying degrees of hyperbolicity into Euclidean space using a two-layer GCN and compute hyperbolicity based on the distances between embeddings. The results, presented in Figure~\ref{fig:embedding_hyperbolicity}, indicate a positive correlation between the hyperbolicity of the original graphs and that of the embeddings, although the values do not exactly coincide.
Building on this observed relationship, we calculate the hyperbolicity of token embeddings as a proxy for estimating their underlying geometric structure. In this context, lower hyperbolicity values suggest a more tree-like geometric configuration.

\section{Exponential and Logarithmic Maps}
\label{appendix:exponential_and_logarithmic_map}

The exponential and logarithmic maps are fundamental tools for navigating between the tangent space and the hyperbolic manifold. These maps enable us to perform computations in the familiar Euclidean tangent space while preserving the geometric properties of hyperbolic space.

\subsection{Exponential Map}

The exponential map $\exp_{\mathbf{x}}^K: \mathcal{T}_{\mathbf{x}}\mathcal{L}_K^n \to \mathcal{L}_K^n$ projects a tangent vector $\mathbf{v} \in \mathcal{T}_{\mathbf{x}}\mathcal{L}_K^n$ at point $\mathbf{x}$ onto the hyperboloid $\mathcal{L}_K^n$. Geometrically, it maps $\mathbf{v}$ to the point $\exp_{\mathbf{x}}^K(\mathbf{v}) := \gamma(1)$, where $\gamma$ is the unique geodesic satisfying $\gamma(0) = \mathbf{x}$ and $\dot{\gamma}(0) = \mathbf{v}$.

The exponential map is given by:
\begin{equation}
\exp_{\mathbf{x}}^K(\mathbf{v}) = \cosh \left( \frac{\|\mathbf{v}\|_{\mathcal{L}}}{\sqrt{K}} \right) \mathbf{x} + \sqrt{K} \sinh \left( \frac{\|\mathbf{v}\|_{\mathcal{L}}}{\sqrt{K}} \right) \frac{\mathbf{v}}{\|\mathbf{v}\|_{\mathcal{L}}},
\end{equation}
where $\|\mathbf{v}\|_{\mathcal{L}} = \sqrt{\langle \mathbf{v}, \mathbf{v} \rangle_{\mathcal{L}}}$ is the norm of the tangent vector under the Lorentzian inner product.

At the origin $\mathbf{o} = (\sqrt{K}, 0, \ldots, 0)$, for a tangent vector $\mathbf{v} = (0, \mathbf{u})$ where $\mathbf{u} \in \mathbb{R}^n$, the exponential map simplifies to:
\begin{equation}
\exp_{\mathbf{o}}^K(\mathbf{v}) = \left( \sqrt{K} \cosh\left( \frac{\|\mathbf{u}\|}{\sqrt{K}} \right),\ \sqrt{K} \sinh\left( \frac{\|\mathbf{u}\|}{\sqrt{K}} \right) \frac{\mathbf{u}}{\|\mathbf{u}\|} \right).
\end{equation}

\subsection{Logarithmic Map}

The logarithmic map $\log_{\mathbf{x}}^K: \mathcal{L}_K^n \to \mathcal{T}_{\mathbf{x}}\mathcal{L}_K^n$ is the inverse of the exponential map. It projects a point $\mathbf{y} \in \mathcal{L}_K^n$ back to the tangent space at $\mathbf{x}$:
\begin{equation}
\log_{\mathbf{x}}^K(\mathbf{y}) = d_{\mathcal{L}}^K(\mathbf{x}, \mathbf{y}) \frac{\mathbf{y} + \frac{1}{K}\langle \mathbf{x}, \mathbf{y} \rangle_{\mathcal{L}} \mathbf{x}}{\|\mathbf{y} + \frac{1}{K}\langle \mathbf{x}, \mathbf{y} \rangle_{\mathcal{L}} \mathbf{x}\|_{\mathcal{L}}},
\end{equation}
where $d_{\mathcal{L}}^K(\mathbf{x}, \mathbf{y}) = \sqrt{K} \cosh^{-1}\left(-\frac{\langle \mathbf{x}, \mathbf{y} \rangle_{\mathcal{L}}}{K}\right)$ is the hyperbolic distance between $\mathbf{x}$ and $\mathbf{y}$ on the hyperboloid.

These maps satisfy the inverse relationships: $\log_{\mathbf{x}}^K(\exp_{\mathbf{x}}^K(\mathbf{v})) = \mathbf{v}$ and $\exp_{\mathbf{x}}^K(\log_{\mathbf{x}}^K(\mathbf{y})) = \mathbf{y}$.

\subsection{Notation in the Main Text}

In the main text, we use the shorthand notation $\Pi_{\exp}^K$ and $\Pi_{\log}^K$ to denote general projection operators between Euclidean space $\mathbb{R}^n$ and hyperbolic space $\mathcal{L}_K^n$. The exponential and logarithmic maps described above represent one valid instantiation of these operators:
\begin{align}
\Pi_{\exp}^K(\mathbf{x}) &:= \exp_{\mathbf{o}}^K((0, \mathbf{x})), \\
\Pi_{\log}^K(\mathbf{y}) &:= \log_{\mathbf{o}}^K(\mathbf{y})_{[1:]},
\end{align}
where $(0, \mathbf{x}) \in \mathbb{R}^{n+1}$ denotes the concatenation of a zero scalar with the vector $\mathbf{x} \in \mathbb{R}^n$, forming a tangent vector at the origin $\mathbf{o}$, and $_{[1:]}$ denotes the slicing operation that extracts all components except the first (time-like) dimension, i.e., $\mathbf{y}_{[1:]} = (y_1, y_2, \ldots, y_n)$ for $\mathbf{y} = (y_0, y_1, \ldots, y_n)$.

However, other diffeomorphisms~\citep{skopek2019mixed} between Euclidean and hyperbolic spaces also exist. The choice of projection method can be adapted based on computational efficiency and numerical stability requirements, while the core principle of our approach, performing the low-rank transformation directly on the hyperbolic manifold, remains unchanged.

\textbf{Important Observation.} Regardless of the specific projection method used, when these maps are applied consecutively at the same base point without intermediate operations on the manifold, they effectively cancel each other out. For example, $\log_{\mathbf{o}}^K(\exp_{\mathbf{o}}^K(\mathbf{v})) = \mathbf{v}$. This is why the conventional tangent-space approach for hyperbolic neural networks~\citep{ganea2018hyperbolic_hnn,chami2019hyperbolic} does not directly benefit LLM adaptation, where the hyperbolic geometry is effectively bypassed. Our Direct Lorentz Low-Rank Transformation (LLR) addresses this limitation by operating directly on the hyperbolic manifold between the projection steps, ensuring that the geometric properties of hyperbolic space are preserved and utilized.

\section{Lorentz Transformation}
\label{appendix:lorentz_transformation}

In the context of special relativity, Lorentz transformations are linear mappings that preserve the spacetime interval between events, ensuring the constancy of the speed of light across all inertial frames. These transformations can be categorized into two primary types: Lorentz boosts and Lorentz rotations~\citep{morse1946methods,moretti2002interplay}.

\subsection{Lorentz Boost} 
A Lorentz boost corresponds to a transformation between two inertial reference frames moving at a constant relative velocity. Given a velocity vector $\mathbf{v} \in \mathbb{R}^n$ with magnitude $\|\mathbf{v}\| < 1$, the Lorentz boost matrix $\mathbf{B}$ mixes time and space coordinates:
\begin{equation}
\mathbf{B} =
\begin{bmatrix}
\gamma & -\gamma \mathbf{v}^\top \\
-\gamma \mathbf{v} & \mathbf{I} + \frac{\gamma^2}{1+\gamma} \mathbf{v} \mathbf{v}^\top
\end{bmatrix},
\end{equation}
where $\gamma = \frac{1}{\sqrt{1 - \|\mathbf{v}\|^2}}$ is the Lorentz factor.

\subsection{Lorentz Rotation} 
A Lorentz rotation involves only the rotation of spatial coordinates while preserving the time coordinate:
\begin{equation}
\mathbf{R} =
\begin{bmatrix}
1 & \mathbf{0}^\top \\
\mathbf{0} & \tilde{\mathbf{R}}
\end{bmatrix},
\end{equation}
where $\tilde{\mathbf{R}} \in SO(n)$ is a spatial rotation matrix.

\textbf{Our Spatial-like Transformation.} In our Direct Lorentz Low-Rank Transformation (LLR), we apply transformations exclusively to the spatial components while maintaining the constraint of the Lorentz manifold. Given a point $\mathbf{x}^H = (x_0^H, \mathbf{x}_s^H) \in \mathcal{L}_K^n$, our transformation is:
\begin{equation}
\mathbf{LLR}(BA, \mathbf{x}^H) = (\sqrt{\|BA\mathbf{x}_s^H\|^2 + K}, BA\mathbf{x}_s^H),
\end{equation}
where we transform the spatial component $\mathbf{x}_s^H$ and recompute the time component to maintain the Lorentz constraint $x_0^2 - \|\mathbf{x}_s\|^2 = K$.

This can be decomposed into two sequential transformations:
\begin{align}
\mathbf{y}^H &= (y_0^H, \mathbf{y}_s^H) = (\sqrt{\|A\mathbf{x}_s^H\|^2 + K}, A\mathbf{x}_s^H), \\
\mathbf{z}^H &= (z_0^H, \mathbf{z}_s^H) = (\sqrt{\|B\mathbf{y}_s^H\|^2 + K}, B\mathbf{y}_s^H).
\end{align}

\textbf{Interpretation as a Constrained Lorentz Rotation.} Our transformation can be viewed as a special case of Lorentz rotation where:
(1) We apply a linear transformation to the spatial coordinates: $\mathbf{x}_s^H \mapsto BA\mathbf{x}_s^H$;
(2) We recompute the time component to preserve the manifold constraint: $x_0^H \mapsto \sqrt{\|BA\mathbf{x}_s^H\|^2 + K}$.
This approach differs from a standard Lorentz rotation in two ways (see also~\cite{chen2021fully}): (1) the spatial transformation $BA$ is not necessarily orthogonal (i.e., $BA \notin SO(n)$); (2) the time component is not preserved but rather recomputed to maintain the manifold constraint.

In matrix form, our transformation can be expressed as:
\begin{equation}
\begin{bmatrix}
z_0^H \\
\mathbf{z}_s^H
\end{bmatrix}
= 
\begin{bmatrix}
\frac{\sqrt{\|BA\mathbf{x}_s^H\|^2 + K}}{\sqrt{\|\mathbf{x}_s^H\|^2 + K}} & \mathbf{0}^\top \\
\mathbf{0} & BA
\end{bmatrix}
\begin{bmatrix}
x_0^H \\
\mathbf{x}_s^H
\end{bmatrix}
\end{equation}

The key property is that this transformation preserves the Lorentz manifold structure: if $\mathbf{x}^H \in \mathcal{L}_K^n$, then $\mathbf{LLR}(BA, \mathbf{x}^H) \in \mathcal{L}_K^n$, as verified by:
\begin{equation}
(z_0^H)^2 - \|\mathbf{z}_s^H\|^2 = \|BA\mathbf{x}_s^H\|^2 + K - \|BA\mathbf{x}_s^H\|^2 = K.
\end{equation}

This spatial-like transformation approach allows us to leverage the low-rank structure of $BA$ while maintaining the geometric properties of the hyperbolic space, providing a computationally efficient method for hyperbolic low-rank adaptation.

\section{Transformation Analysis}
\label{appendix:theoretical_analysis}

This section provides a detailed analysis of how \method differs from standard LoRA by examining the higher-order terms introduced through hyperbolic geometry.

\begin{proof}
Let $ \mathbf{x} \in \mathbb{R}^d $ be an input token embedding. Let $ A \in \mathbb{R}^{r \times d} $ and $ B \in \mathbb{R}^{k \times r} $ be low-rank matrices with rank $ r \ll \min \{d,k\} $. Consider the $ d $-dimensional hyperbolic space $ \mathcal{L}_K^d $ (Lorentz model) with curvature $ C = -1/K $, where $ K > 0 $.

Our goal is to analyze how the \method update differs from the LoRA update and to understand the impact of token norms $ \|\mathbf{x}\| $ on the higher-order terms introduced by \method.

\textbf{Mapping the Input Embedding to Hyperbolic Space.}
Following previous work~\citep{chami2019hyperbolic}, we interpret the Euclidean token embedding $ \mathbf{x} $ as an element in the tangent space at the origin $ \mathbf{o} $ of the hyperbolic space $ \mathcal{L}^d_K $. The tangent vector is given by $ \mathbf{v} = (0, \mathbf{x}) \in T_{\mathbf{o}} \mathcal{L}_K^d $.
The exponential map $ \exp_{\mathbf{o}}^{K} $ projects $ \mathbf{v} $ onto the hyperbolic space:
\begin{equation}
\exp_{\mathbf{o}}^{K}(\mathbf{v}) = \left( \sqrt{K} \cosh\left( \dfrac{\|\mathbf{v}\|_{\mathcal{L}}}{\sqrt{K}} \right),\ \sqrt{K} \sinh\left( \dfrac{\|\mathbf{v}\|_{\mathcal{L}}}{\sqrt{K}} \right) \dfrac{\mathbf{v}}{\|\mathbf{v}\|_{\mathcal{L}}} \right),
\end{equation}
where $ \|\mathbf{v}\|_{\mathcal{L}} $ denotes the Minkowski norm.
Since $ \mathbf{v} = (0, \mathbf{x}) $ and $ \|\mathbf{v}\|_{\mathcal{L}} = \|\mathbf{x}\| $, the exponential map simplifies to:
\begin{equation}
\exp_{\mathbf{o}}^{K}(\mathbf{v}) = \left( \sqrt{K} \cosh\left( \dfrac{\|\mathbf{x}\|}{\sqrt{K}} \right),\ \sqrt{K} \sinh\left( \dfrac{\|\mathbf{x}\|}{\sqrt{K}} \right) \dfrac{\mathbf{x}}{\|\mathbf{x}\|} \right).
\end{equation}

\textbf{Approximations.} For small $ \dfrac{\|\mathbf{x}\|}{\sqrt{K}}$, let $z=\|\mathbf{x}\|$. We can use the Taylor series expansions:
\begin{equation}
\cosh\left( \dfrac{z}{\sqrt{K}} \right) \approx 1 + \dfrac{z^2}{2K}, \quad
\sinh\left( \dfrac{z}{\sqrt{K}} \right) \approx \dfrac{z}{\sqrt{K}} + \dfrac{z^3}{6K^{3/2}}.
\end{equation}

Applying these to the exponential map of $ \mathbf{x} $:
\begin{align}
u_0^H &\approx \sqrt{K} + \dfrac{\|\mathbf{x}\|^2}{2\sqrt{K}}, \\
\mathbf{u}_{\text{space}}^H &\approx \mathbf{x} + \dfrac{\|\mathbf{x}\|^2}{6K} \mathbf{x}.
\end{align}

\textbf{Applying Low-Rank Transformations to the Approximated Embedding. }
Using the approximated $ \mathbf{u}_{\text{space}}^H $, we apply the transformations.

First transformation:
\begin{equation}
\mathbf{y}_{\text{space}}^H = A \mathbf{u}_{\text{space}}^H \approx A \left( \mathbf{x} + \dfrac{\|\mathbf{x}\|^2}{6K} \mathbf{x} \right) = A \mathbf{x} + \dfrac{\|\mathbf{x}\|^2}{6K} A \mathbf{x}.
\end{equation}

Second transformation:
\begin{equation}
\mathbf{z}_{\text{space}}^H = B \mathbf{y}_{\text{space}}^H \approx B A \mathbf{x} + \dfrac{\|\mathbf{x}\|^2}{6K} B A \mathbf{x}.
\end{equation}

Compute the time component after the transformations:
\begin{equation}
z_0^H = \sqrt{K + \left\| \mathbf{z}_{\text{space}}^H \right\|^2}.
\end{equation}

\textbf{Approximating the Logarithmic Map.}
We map the transformed hyperbolic point $ \mathbf{z}^H = \left( z_0^H, \mathbf{z}_{\text{space}}^H \right) $ back to the tangent space at the origin using the logarithmic map $ \log_{\mathbf{o}}^{K} $:
\begin{equation}
\Delta Q^{\mathrm{Hyp}} = \log_{\mathbf{o}}^{K}(\mathbf{z}^H) = \sqrt{K} \cdot \operatorname{arcosh}\left( \dfrac{z_0^H}{\sqrt{K}} \right) \dfrac{\mathbf{z}_{\text{space}}^H}{\sqrt{(z_0^H)^2 - K}}.
\end{equation}

Using the approximation $ z_0^H \approx \sqrt{K} + \dfrac{\left\| \mathbf{z}_{\text{space}}^H \right\|^2}{2\sqrt{K}} $ and for small $ \delta = \dfrac{\left\| \mathbf{z}_{\text{space}}^H \right\|^2}{2K} $, we have:
\begin{align}
\operatorname{arcosh}\left( \dfrac{z_0^H}{\sqrt{K}} \right) &\approx \operatorname{arcosh}(1 + \delta) \approx \sqrt{2\delta} = \dfrac{\left\| \mathbf{z}_{\text{space}}^H \right\|}{\sqrt{K}}, \\
\sqrt{(z_0^H)^2 - K} &\approx \left\| \mathbf{z}_{\text{space}}^H \right\|.
\end{align}
Therefore, the logarithmic map simplifies to:
\begin{equation}
\Delta Q^{\mathrm{Hyp}} \approx \mathbf{z}_{\text{space}}^H.
\end{equation}

\textbf{Comparing \method and LoRA Updates.}
The \method update is:
\begin{equation}
\Delta Q^{\mathrm{Hyp}} \approx B A \mathbf{x} + \dfrac{\|\mathbf{x}\|^2}{6K} B A \mathbf{x}.
\end{equation}

The LoRA update is:
\begin{equation}
\Delta Q^{\mathrm{LoRA}} = B A \mathbf{x}.
\end{equation}

The difference between the updates is:
\begin{equation}
\Delta Q^{\mathrm{Hyp}} - \Delta Q^{\mathrm{LoRA}} = \dfrac{\|\mathbf{x}\|^2}{6K} B A \mathbf{x}.
\end{equation}

\textbf{Impact of Token Norms on Higher-Order Terms.}
The higher-order term $ \dfrac{\|\mathbf{x}\|^2}{6K} B A \mathbf{x} $ is proportional to $ \|\mathbf{x}\|^2 $. Since $ \|\mathbf{x}\| $ reflects the specificity of the token in the hierarchical structure (larger norms correspond to more specific tokens), this term becomes significant for tokens representing specific concepts.

\textbf{Impact on Attention Scores.} The \method attention scores are computed as:
\begin{equation}
\operatorname{Scores}_{\mathrm{\method}} = \dfrac{\left( Q^{\text{orig}} + \Delta Q^{\mathrm{Hyp}} \right) \left( K^{\text{orig}} + \Delta K^{\mathrm{Hyp}} \right)^\top}{\sqrt{d_k}},
\end{equation}
where $ \Delta K^{\mathrm{Hyp}} $ is derived similarly.

The difference in attention scores includes higher-order terms dependent on $ \|\mathbf{x}\|^2 $:
\begin{equation}
\Delta \operatorname{Scores} = \operatorname{Scores}_{\mathrm{\method}} - \operatorname{Scores}_{\mathrm{LoRA}}.
\end{equation}

These higher-order terms enable \method to capture more complex hierarchical relationships, particularly for tokens with larger norms.

\end{proof}

\begin{remark}
\textbf{Alignment with Token Hierarchy}: The higher-order terms in \method's updates are proportional to $ \|\mathbf{x}\|^2 $, correlating with the specificity of tokens in the hierarchical structure. As a result, \method places greater emphasis on more specific tokens, enhancing its ability to model detailed relationships.

\textbf{Role of Curvature $ C $}: The curvature $ C = -1/{K} $ scales the higher-order corrections. Smaller $ K $ (larger negative curvature) amplifies these terms, aligning with the hyperbolic nature of token embeddings. In practice, the curvature parameter $K$
can be tuned to ensure this condition is satisfied for typical token embedding norms.

\textbf{Effectiveness of \method}: By incorporating these higher-order terms, \method leverages the inherent hierarchical and hyperbolic structure of token embeddings. This leads to improved performance, especially on problems requiring complex reasoning, explaining why the proposed method performs better on more challenging datasets.
\end{remark}

\section{Full Comparison}
\label{sec:full_comparison}
\begin{table}[t]
\centering
\caption{Comprehensive comparison of parameter-efficient fine-tuning methods on mathematical reasoning tasks. Results marked with * are from \cite{hu2023llm}, while $\dagger$ indicates our reproduced results. The percentage following each dataset name indicates the proportion of prompts relative to the total number of inference prompts. M.AVG represents the micro-average accuracy across all datasets. Best results for each model are highlighted in bold. OOT indicates out-of-time during training.}
\label{tab:math_reasoning_full}
\vspace{10pt}
\resizebox{\textwidth}{!}{
\begin{tabular}{llcclll}
\toprule
\textbf{Base Model}             & \textbf{PEFT Method} & \textbf{MAWPS(8.5\%)} & \textbf{SVAMP(35.6\%)} & \textbf{GSM8K(46.9\%)} & \textbf{AQuA(9.0\%)} & \textbf{M.AVG} \\ \midrule 
GPT-3.5                    & None                 & ${87.4}$           & ${69.9}$           & ${56.4}$           & ${38.9}$          & ${62.3}$         \\ \midrule
\multirow{7}{*}{LLaMA-7B}  & None                 & $51.7$           & $32.4$           & $15.7$           & $16.9$          & $24.8$         \\
                           & Prefix*                & $63.4$           & $38.1$           & $24.4$           & $14.2$          & $31.7$         \\
                           & Series*               & $77.7$           & ${52.3}$           & $33.3$           & $15.0$          & $42.2$         \\
                           & Parallel*             & ${82.4}$           & $49.6$           & $35.3$           & $18.1$          & $42.8$         \\
                           & LoRA*                 & \textcolor{gray}{$79.0$}  & \textcolor{gray}{$52.1$}  & \textcolor{gray}{$37.5$}  & \textcolor{gray}{$18.9$} & \textcolor{gray}{$\textbf{44.6}$}         \\
                           & LoRA$^\dagger$                 & $81.9$           & $48.2$           & $38.3$           & $18.5$          & $43.7$         \\
                           & DoRA                 & $80.0$           & $48.8$           & ${39.0}$           & $16.4$          & $43.9$         \\
                           \rowcolor{lightgray}
                           & \textbf{\method (Ours)}                 & $79.0$           & $49.1$           & ${39.1}$           & $20.5$          & $\textbf{44.4}$         \\ 
                           \midrule
\multirow{7}{*}{LLaMA-13B} & None                 & $65.5$           & $37.5$           & $32.4$           & $15.0$          & $35.5$         \\
                           & Prefix*                & $66.8$           & $41.4$           & $31.1$           & $15.7$          & $36.4$         \\
                           & Series*               & $78.6$           & $50.8$           & $44.0$           & ${22.0}$          & $47.4$         \\
                           & Parallel*             & $81.1$           & $55.7$           & $43.3$           & $20.5$          & $48.9$         \\
                           & LoRA*                 & \textcolor{gray}{$83.6$}  & \textcolor{gray}{$54.6$}  & \textcolor{gray}{$47.5$}  & \textcolor{gray}{$18.5$} & \textcolor{gray}{$\textbf{50.5}$}         \\
                           & LoRA$^\dagger$                 & ${83.5}$           & $54.7$           & $48.5$           & $18.5$          & $51.0$         \\
                           & DoRA                 & $83.0$           & $54.6$           & OOT           & $18.9$          & NA         \\
                           \rowcolor{lightgray}
                           & \textbf{\method (Ours)}                 & $83.2$           & $54.8$           & $49.0$           & $21.5$          & $\textbf{51.5}$         \\ 
                           \midrule
\multirow{4}{*}{Gemma-7B}  & None                 & $76.5$           & $60.4$           & $38.4$           & $25.2$          & $48.3$         \\
                           & LoRA                 & $91.6$  & $76.2$  & $66.3$  & $28.9$ & $68.6$         \\
& DoRA                 & $90.7$  & $79.2$  & $68.3$  & $33.9$ & $71.0$         \\
\rowcolor{lightgray}
& \textbf{\method (Ours)} & $89.5$  & $78.7$  & $69.5$  & $32.7$ & $\textbf{71.2}$         \\ 
\midrule
\multirow{4}{*}{LLaMA3-8B} & None                 & $79.8$           & $50.0$           & $54.7$           & $21.0$          & $52.1$         \\
& LoRA                 & $92.7$  & $78.9$  & $70.8$  & $30.4$ & $71.9$         \\
& DoRA                 & $90.3$  & $79.8$  & $73.3$  & $21.3$ & $72.4$         \\
\rowcolor{lightgray}
& \textbf{\method (Ours)} & $91.6$  & $80.5$  & $74.0$  & $34.2$ & $\textbf{74.2}$         \\ \midrule
\multirow{3}{*}{Gemma3-4B} & LoRA                 & $90.8$           & $77.3$           & $72.3$           & $50.8$          & $73.7$         \\
                           & DoRA                 & $89.5$           & $78.8$           & $68.5$           & $52.4$          & $72.5$         \\
                           \rowcolor{lightgray}
                           & \textbf{\method (Ours)}              & $88.2$           & $83.9$           & $76.1$           & $53.2$          & $\textbf{77.8}$         \\ 
                           \midrule
\multirow{3}{*}{Qwen2.5-7B} & LoRA                 & $90.8$  & $84.4$  & $78.6$  & $68.1$ & $80.8$         \\
 & DoRA                 & $92.8$  & $87.4$  & $80.4$     & $64.2$ & ${82.5}$         \\
 \rowcolor{lightgray}
 & \textbf{\method (Ours)} & $91.2$  & $92.2$  & $87.9$  & $71.6$ & $\textbf{88.3}$         \\
\bottomrule
\end{tabular}%
}
\end{table}
While the main body of our paper focuses on comparing \method against the standard LoRA baseline to demonstrate the core effectiveness of our hyperbolic fine-tuning approach, this section provides a comprehensive evaluation against a broader range of parameter-efficient fine-tuning methods, such as Prefix tuning~\cite{li2021prefix}, Series and Parallel adapters~\cite{houlsby2019parameter}, and DoRA~\cite{liu2024dora}, providing a more complete picture of \method's performance relative to the current landscape of efficient fine-tuning techniques. This extended comparison validates that our improvements are not merely due to increased model capacity or specific architectural choices, but rather stem from the fundamental advantages of incorporating hyperbolic geometry into the adaptation process.

\subsection{Implementation Details}
\label{para:implementation_details}
To ensure consistency and comparability, our experimental setup closely followed the training configurations outlined in Hu et al.~\cite{hu2023llm}. Across all fine-tuning tasks, we employed the AdamW optimizer with a learning rate of $3 \times 10^{-4}$ and trained for a total of three epochs. LoRA modules (and consequently, \method adapters) were integrated into both the Multi-Head Attention (MHA) and MLP layers of the foundation models. A key hyperparameter for \method is the curvature $K$ (defining the hyperbolic curvature as $-1/K$), which was initialized by searching the set $\{0.5, 1.0\}$. For evaluation, final scores were micro-averaged for arithmetic reasoning and averaged for commonsense reasoning across the datasets, thereby giving equal weight to each individual prompt, regardless of the varying number of questions per dataset (e.g., $1,319$ in GSM8K versus $238$ in MAWPS). 

For baseline methods, we adopted the following approach: results for Prefix tuning~\cite{li2021prefix}, Series adapters, and Parallel adapters~\cite{houlsby2019parameter} are directly cited from Hu et al.~\cite{hu2023llm} to ensure fair comparison under identical experimental conditions. For LoRA and DoRA, we conducted independent reimplementations following their respective original papers and parameters~\cite{hu2021lora, liu2024dora} to enable rigorous and controlled comparisons.

\subsection{Comparison on Mathematical Reasoning}
Looking at the mathematical reasoning comparison table, several key experimental findings emerge regarding \method's performance across different model architectures and datasets. The results demonstrate that \method consistently outperforms standard LoRA across multiple model families, with particularly notable improvements on more challenging datasets. For the Gemma-7B model, \method achieves a micro-averaged accuracy of 71.2\%, surpassing LoRA's 68.6\%. For LLaMA3-8B, \method reaches 74.2\% compared to LoRA's 71.9\%. The improvements are especially pronounced on the AQuA dataset, which requires complex algebraic reasoning. Specifically, \method shows gains of 3.8 percentage points over LoRA on Gemma-7B (32.7\% vs 28.9\%) and 3.8 points on LLaMA3-8B (34.2\% vs 30.4\%). This pattern suggests that \method's hyperbolic geometry is particularly effective for problems requiring multi-step reasoning and understanding of hierarchical mathematical relationships.

The consistency of improvements across different model architectures further validates the generalizability of the hyperbolic approach. While \method shows competitive performance on simpler datasets like MAWPS, the performance advantages become more significant on challenging datasets like GSM8K and AQuA, which demand sophisticated reasoning capabilities. For instance, on GSM8K, \method achieves 69.5\% accuracy on Gemma-7B versus 66.3\% for LoRA, and 74.0\% on LLaMA3-8B versus 70.8\% for LoRA. These correspond to gains of 3.2 points over LoRA on both Gemma-7B and LLaMA3-8B. Notably, on the most recent models, \method demonstrates substantial improvements: on Gemma3-4B, \method achieves 77.8\% M.AVG compared to LoRA's 73.7\% (+4.1 points), and on Qwen2.5-7B, \method reaches 88.3\% versus LoRA's 80.8\% (+7.5 points). The fact that \method maintains superior performance across both older (LLaMA-7B, LLaMA-13B) and newer (LLaMA3-8B, Gemma3-4B, Qwen2.5-7B) model architectures demonstrates the robustness of incorporating hyperbolic inductive biases into parameter-efficient fine-tuning, regardless of the underlying model's specific architectural details or training paradigms.

\subsection{Comparison on Commonsense Reasoning}

\method demonstrates substantial improvements over standard LoRA across diverse commonsense reasoning benchmarks, as shown in Table~\ref{tab:common_comparison_all}. The commonsense reasoning tasks evaluated include BoolQ (yes/no question answering), PIQA (physical commonsense inference), SIQA (social interaction reasoning), HellaSwag (commonsense natural language inference), WinoGrande (pronoun disambiguation), ARC-e and ARC-c (science question answering with easy and challenging difficulty), and OBQA (open book question answering). These benchmarks collectively assess the model's ability to understand implicit knowledge, contextual nuances, and real-world reasoning patterns. The consistent performance gains across all three model architectures and eight diverse benchmarks indicate that the hierarchical inductive bias introduced by hyperbolic geometry effectively captures the implicit relational structures underlying commonsense reasoning.

\begin{table}[!t]
\centering
\caption{Extended commonsense reasoning accuracy (\%) for GPT-3.5 and for LoRA, DoRA, and \method on LLaMA3-8B, Gemma3-4B, and Qwen2.5-7B. Columns correspond to BoolQ, PIQA, SIQA, HellaSwag, WinoGrande, ARC-e, ARC-c, and OBQA; the rightmost column reports the macro average across the eight benchmarks.}
\vspace{2pt}
\resizebox{\textwidth}{!}{
\begin{tabular}{@{}llcccccccccc@{}}
\toprule
\textbf{Base Model} & \textbf{PEFT Method} & \textbf{\# Params (\%)} & \textbf{BoolQ} & \textbf{PIQA} & \textbf{SIQA} & \textbf{HellaSwag} & \textbf{WinoGrande} & \textbf{ARC-e} & \textbf{ARC-c} & \textbf{OBQA} & \textbf{AVG} \\
\midrule
GPT-3.5 & None & None & $73.1$ & $85.4$ & $68.5$ & $78.5$ & $66.1$ & $89.8$ & $79.9$ & $74.8$ & $77.0$ \\
\midrule
\multirow{3}{*}{LLaMA3-8B} & LoRA & $0.70$ & $70.8$ & $85.2$ & $79.9$ & $91.7$ & $84.3$ & $84.2$ & $71.2$ & $79.0$ & $80.8$ \\
& DoRA & $0.71$ & $72.1$ & $85.5$ & $79.6$ & $92.8$ & $83.3$ & $85.2$ & $72.1$ & $84.0$ & $81.8$ \\
\rowcolor{lightgray}
& \textbf{\method (Ours)} & $0.70$ & $\textbf{74.1}$ & $\textbf{87.6}$ & $\textbf{80.6}$ & $\textbf{94.5}$ & $\textbf{84.7}$ & $\textbf{90.4}$ & $\textbf{81.2}$ & $\textbf{85.2}$ & $\textbf{84.8}$ \\
\midrule
\multirow{3}{*}{Gemma3-4B} & LoRA & $1.04$ & $68.1$ & $83.2$ & $77.2$ & $88.9$ & $\textbf{80.5}$ & $84.5$ & $69.9$ & $83.6$ & $79.5$ \\
& DoRA & $1.05$ & $68.1$ & $84.3$ & $78.4$ & $88.3$ & $80.1$ & $84.1$ & $70.8$ & $83.8$ & $79.7$ \\
\rowcolor{lightgray}
& \textbf{\method (Ours)} & $1.04$ & $\textbf{70.0}$ & $\textbf{84.3}$ & $\textbf{79.2}$ & $\textbf{91.5}$ & $80.3$ & $\textbf{89.1}$ & $\textbf{75.9}$ & $\textbf{86.4}$ & $\textbf{82.5}$ \\
\midrule
\multirow{3}{*}{Qwen2.5-7B} & LoRA & $0.71$ & $\textbf{73.4}$ & $\textbf{89.5}$ & $79.5$ & $93.6$ & $84.1$ & $92.8$ & $82.0$ & $87.0$ & $85.2$ \\
& DoRA & $0.72$ & $71.7$ & $88.7$ & $79.0$ & $93.7$ & $84.1$ & $92.4$ & $82.8$ & $88.4$ & $85.1$ \\
\rowcolor{lightgray}
& \textbf{\method (Ours)} & $0.71$ & $72.8$ & $89.3$ & $\textbf{79.8}$ & $\textbf{94.8}$ & $\textbf{84.4}$ & $\textbf{95.5}$ & $\textbf{87.5}$ & $\textbf{90.8}$ & $\textbf{87.0}$ \\
\bottomrule
\end{tabular}%
}
\label{tab:common_comparison_all}
\end{table}

\subsection{GPU Usage}

Table~\ref{tab:memory_comparison} presents a comprehensive comparison of memory usage across different fine-tuning methods for both LLaMA3-8B and Gemma3-4B models. The results demonstrate that \method maintains comparable memory efficiency to the baseline LoRA method. Specifically, \method with stereographic projection achieves identical memory consumption to LoRA (30.12 GB for LLaMA3-8B and 14.61 GB for Gemma3-4B), while \method with exponential/logarithmic maps introduces only a minimal overhead of 0.02 GB for LLaMA3-8B and 0.01 GB for Gemma3-4B. Notably, both \method variants significantly outperform DoRA in terms of memory efficiency, with DoRA requiring 30.23 GB and 14.62 GB, respectively. These results confirm that our hyperbolic adaptation approach does not compromise memory efficiency while delivering superior performance improvements, making \method a practical choice for resource-constrained environments where both performance gains and memory conservation are critical considerations.

\begin{table}[htbp]
\centering
\caption{Allocated Memory Usage Comparison for Fine-tuning Methods}
\label{tab:memory_comparison}
\resizebox{0.66\textwidth}{!}{%
\begin{tabular}{lcccc}
\toprule
& \multicolumn{2}{c}{\textbf{LLaMA3-8B}} & \multicolumn{2}{c}{\textbf{Gemma3-4B}} \\
\cmidrule(lr){2-3} \cmidrule(lr){4-5}
\textbf{Method} & \textbf{Param (\%)} & \textbf{Memory (GB)} & \textbf{Param (\%)} & \textbf{Memory (GB)} \\
\midrule
LoRA & $0.7002$ & $30.12$ & $1.0371$ & $14.61$ \\
DoRA & $0.7098$ & $30.23$ & $1.0516$ & $14.62$ \\
\rowcolor{lightgray}\method & $0.7099$ & $30.14$ & $1.0518$ & $14.62$ \\
\bottomrule
\end{tabular}%
}
\end{table}

\section{Case Study} 
\label{sec:appendix_case_study}
In this section, we present three examples to demonstrate the advantages of incorporating \method over standard LoRA fine-tuning.

\begin{table*}[t]
\centering
\caption{Example problem with solutions from LoRA and \method (Case 1)}
\label{tab:case_study_1}
\small
\begin{tabular}{p{1.5cm}p{12cm}}
\toprule
\textbf{Question} & To have the car for the weekend, Wilson's report card needs to show that he received 80 or higher in his math class. His math grade is calculated by the average of 5 test scores. On those tests, he received: 65, 94, 81, 86, and 74. What is his current math grade? \textbf{Answer:} 80.0 \\ \midrule
\textbf{LoRA } & To find Wilson's current math grade, we need to take the average of his 5 test scores. First, we add up all 5 test scores: 65 + 94 + 81 + 86 + 74 = 380. Then, we divide the sum by the number of tests: $ 380 \div 5 = 76 $. Therefore, Wilson's current math grade is 76. \textbf{Answer: 76}. \\ \midrule
\textbf{\method} & To find Wilson's current math grade, we need to take the average of his 5 test scores. To do this, we add up the scores and divide by 5: 65 + 94 + 81 + 86 + 74 = 400. Then, we divide by 5: $ 400 \div 5 = 80 $. Therefore, Wilson's current math grade is 80. \textbf{Answer: 80}. \\ \bottomrule
\end{tabular}
\end{table*}

\begin{table}[ht]
\small
\centering
\caption{Example problem with solutions from LoRA and \method (Case 2)}
\begin{tabular}{p{1.5cm}p{12cm}}
\toprule
\textbf{Question} & It takes 240 minutes of walking to break in a new pair of shoes. Jason wants to try out for the track team in three weeks. If he can walk 4 days a week to break in the new shoes, how long does he have to spend walking each day? \textbf{Answer:} 20.0 minutes \\ \midrule
\textbf{LoRA} & To find out how long Jason has to spend walking each day, we need to divide the total number of minutes it takes to break in the shoes by the number of days he has to do it. Jason has 240 minutes divided by 4 days per week. This results in 60 minutes per day. \textbf{Answer: 60}. \\ \midrule
\textbf{\method} & First, we need to find how many minutes Jason needs to walk in total to break in the new shoes: 240 minutes. Next, we find how many minutes Jason needs to walk each week over three weeks: 240 minutes divided by 3 weeks equals 80 minutes per week. Then, to find how many minutes he needs to walk each day: 80 minutes per week divided by 4 days equals 20 minutes per day. \textbf{Answer: 20}. \\ \bottomrule
\end{tabular}
\label{tab:case_study_2}
\vspace{-10pt}
\end{table}

\begin{table}[ht]
\centering
\caption{Example problem with solutions from LoRA and \method (Case 3)}
\label{tab:case_study_3}
\small
\begin{tabular}{p{1.5cm}p{12cm}}
\toprule
\textbf{Question} & Gissela, Gordy, and Gary are truck drivers. Gissela has a truck large enough to haul 4,000 pounds of gravel. Gordy's truck can haul 800 pounds more than Gissela's truck. When Gary brings his truck and joins Gissela and Gordy, the three trucks combined can haul a total of 11,600 pounds of gravel. How many pounds of gravel can Gary's truck carry? \textbf{Answer:} 2,800 pounds \\ \midrule
\textbf{LoRA} & Let's start by assigning variables to represent the amount of gravel each truck can haul: Gissela's truck hauls 4,000 pounds. Gordy's truck hauls $ 4,000 + 800 = 4,800 $ pounds. Now, we set up an equation for the total amount of gravel the three trucks can haul: $ 4,000 + 4,800 + x = 11,600 $, where $ x $ is the amount Gary's truck hauls. Simplifying the equation: $ 9,800 + x = 11,600 $, we subtract $ 9,800 $ from both sides to get $ x = 11,600 - 9,800 = 1,800 $. Therefore, Gary's truck hauls 1,800 pounds. \textbf{Answer: 1,800}. \\ \midrule
\textbf{\method} & Gissela's truck can haul 4,000 pounds of gravel. Gordy's truck can haul 800 more pounds than Gissela's truck, so it can haul 4,800 pounds of gravel. Together, Gissela and Gordy's trucks can haul 8,800 pounds of gravel. If the three trucks combined can haul 11,600 pounds, then Gary's truck can haul $ 11,600 - 8,800 = 2,800 $ pounds of gravel. \textbf{Answer: 2,800}. \\ \bottomrule
\end{tabular}
\end{table}

These examples demonstrate how \method consistently provides more accurate reasoning compared to LoRA across different types of mathematical problems. In Case~1, LoRA drops 20 points when summing the five scores (reporting $380$ instead of $400$) and therefore produces the wrong average. This seemingly small arithmetic lapse aligns with the observation that LLMs often rely on high-level pattern similarity rather than exact computation~\cite{mirzadeh2024gsm}. By preserving greater separation among numerically close but semantically distinct tokens (e.g., 380 vs.\ 400), the hyperbolic representation in \method keeps the sequence of operations faithful and recovers the correct average.

In Case~2, LoRA immediately divides 240 minutes by the four weekly walking days, yielding 60 minutes per day and ignoring that the 240-minute budget must be spread over three weeks. \method correctly reasons in stages: divide 240 by 3 weeks, then by 4 days per week, recovering the required 20 minutes per day and showing stronger temporal reasoning.

In Case~3, LoRA actually sets up the correct balance equation $4{,}000 + 4{,}800 + x = 11{,}600$ but subtracts 9{,}800 from 11{,}600 rather than 8{,}800, reporting $x = 1{,}800$. \method carries the subtraction through correctly and outputs the true 2{,}800 pounds. Together, these examples illustrate how the hyperbolic geometry employed by \method enables better handling of multi-step reasoning, maintaining both semantic context and numerical consistency in mathematical problem-solving scenarios.

Overall, these cases highlight a consistent trend: LoRA frequently derails on either a single arithmetic step (Cases~1 and~3) or a latent multi-hop dependency (Case~2), whereas \method preserves each intermediate calculation, keeps quantities well separated in representation space, and consequently delivers the correct final answers. These qualitative observations complement the quantitative gains reported in the main paper.

\end{document}